\documentclass[10pt,conference,letterpaper]{IEEEtran}
\IEEEoverridecommandlockouts
\usepackage[english]{babel}
\usepackage{cite}
\usepackage{algorithmic}
\usepackage{textcomp}

\usepackage[utf8]{inputenc} 
\usepackage[T1]{fontenc}    
\usepackage[colorlinks=false,linkcolor=black,citecolor=black,anchorcolor=black,pdfborder={0 0 0}]{hyperref}
\usepackage{url}            
\usepackage{booktabs}       
\usepackage{amsfonts}       
\usepackage{nicefrac}       
\usepackage[kerning,spacing]{microtype}      
\usepackage{xcolor}         

\usepackage{graphicx}
\usepackage[font=small,labelfont=bf]{caption}
\usepackage{subcaption}

\usepackage{amsmath}
\usepackage{amssymb}
\usepackage{mathtools}
\usepackage{amsthm}

\usepackage{enumitem}

\usepackage{multirow}
\usepackage{ulem}

\usepackage{xspace}

\usepackage{mathptmx}
\DeclareMathAlphabet{\mathcal}{OMS}{cmsy}{m}{n}
\DeclareMathAlphabet{\mathrm}{OT1}{bch}{m}{n}
\DeclareMathAlphabet{\mathit}{OT1}{bch}{m}{it}

\newcommand{\modelname}{\textrm{RAVEN}\xspace}
\newcommand{\para}[1]{\smallskip\noindent\textbf{#1}}

\def\BibTeX{{\rm B\kern-.05em{\sc i\kern-.025em b}\kern-.08em
    T\kern-.1667em\lower.7ex\hbox{E}\kern-.125emX}}
\begin{document}

\title{RAVEN: A Regime-Aware Variable-context Expert Network for Financial Time Series Forecasting\\

\thanks{*Corresponding authors}
}

\author{
    \IEEEauthorblockN{
        Cheng He$^{1,2}$, 
        Zhenyu Guan$^{2}$, 
        Xijie Liang$^{2}$, 
        Defu Lian$^{1*}$, 
        Jiajia Li$^{3*}$,
        Enhong Chen$^{1}$,\\ 
        Patrick P. C. Lee$^{4}$, 
        Geng Hu$^{5}$, 
        Zehao Chen$^{2}$
    }
    \IEEEauthorblockA{
        $^{1}$University of Science and Technology of China, China \ \
        $^{2}$Shanghai Black Wing Asset Management Co., Ltd., China\\
        $^{3}$College of Sciences, Shanghai University, China\ \
        $^{4}$The Chinese University of Hong Kong, China\\
        $^{5}$Qitan Technology Co., Ltd., China
    }
}

\maketitle

\begin{abstract}
Financial time series forecasting presents structural challenges absent from standard benchmarks. Log-returns are non-stationary, exhibit exceptionally low signal-to-noise (SNR) ratios, and are governed by regime-dependent temporal dependencies.
We identify a key limitation of state-of-the-art (SOTA) time series models in financial settings. A fixed context window is mismatched to the time-varying optimal look-back of non-stationary price processes.
We propose the Regime-Aware Variable-context Expert Network (RAVEN), a Mixture-of-Experts framework designed to adaptively determine the temporal context for each input sample. Instead of relying on a fixed look-back horizon, RAVEN constructs a hierarchy of nested contiguous windows whose lengths are determined by the data itself. 
Specifically, \modelname scores patches by learned importance in reverse chronological order and applies the Cumulative Importance Thresholding (CIT) mechanism to derive nested prefix windows, each routed to a scale-specialized expert.
A Global Compressed Representation (GCR) branch runs in parallel over the full context, preserving global temporal coherence that local experts cannot guarantee. 
Because the nested routing induces structured overlap among expert inputs, we introduce a Correlation-Aware Weighting (CAW) to align variable-length expert outputs and penalize pairwise cosine similarity prior to aggregation. 
Experiments on cumulative log-return prediction (HS300, S\&P500) and fund sales forecasting demonstrate that \modelname achieves SOTA performances, improves Pearson correlation by 9.2\% on HS300 and 20.2\% on S\&P500, and reduces MSE by 18.2\% on fund sales forecasting, while achieving the best results in 14 of 16 metrics on four PEMS traffic benchmarks.
\end{abstract}


\section{Introduction}
Financial time series forecasting is a cornerstone of quantitative investment, supporting tasks from risk management to automated trading. Unlike well-known benchmark datasets in the general time series domain, such as ETTh, ETTm, Weather, Electricity, or Traffic \cite{Wu21}, which often exhibit clear periodic patterns and deterministic trends, financial data is noisy and non-stationary \cite{lo2004}. In this adversarial environment, predicting raw price levels $C_t$ at time $t$ is impractical: prices are not scale-invariant and cross-asset comparable, typically follow a random walk, and exhibit extremely high autocorrelation, leading to spurious regression and inflated in-sample performance. To capture meaningful predictive signals, existing approaches reformulate the problem as the regression of log-returns: $r_t = \ln(C_t / C_{t-1})$~\cite{hasbrouck91, gu20}.
This transformation shifts the modeling objective from tracking absolute values to capturing price innovations \cite{hasbrouck91}, the unpredictable stochastic components driven by the arrival of new market information. 
While log-returns are statistically more stable than raw prices, they still exhibit exceptionally low SNR and heavy-tailed distributions, making accurate regression a challenging task~\cite{gu20}.

Historically, financial time-series analysis has been dominated by tree-based ensembles such as XGBoost~\cite{chen16} and LightGBM~\cite{ke17}, descendants of the gradient boosting framework~\cite{friedman01}. They excel at modeling non-linear interactions over handcrafted technical features,  but treat each forecast as a static tabular problem and ignore the inherent temporal topology of market regimes. Subsequent deep-learning architectures, from Multi-Layer Perceptron through Recurrent Neural Networks~\cite{elman90} to Long Short-Term Memory networks~\cite{hochreiter97} and Gated Recurrent Unit~\cite{Cho14}, restore a notion of temporal memory through recurrence and gating, yet remain biased toward recent observations and often fail to distinguish transient market noise from long-term structural regime shifts.

Recently, Transformer-based models, from Informer~\cite{Zhou21}, PatchTST~\cite{Nie23}, TimesNet~\cite{Wu23}, and iTransformer~\cite{Liu24} to frequency-domain variants such as FredFormer~\cite{Piao24}, and WPMixer~\cite{Murad25}, have enhanced general time-series forecasting by leveraging attention to capture long-range dependencies. However, when applied to financial markets, these models inherit a structural bottleneck that is rarely examined: \textbf{the reliance on a fixed-length historical context window $L$}. In non-stationary financial environments, a static $L$ creates an irreconcilable conflict: a short window lacks memory to span structural regime shifts, while a long window unavoidably mixes stale information from a prior regime into the current prediction as additive noise.

Classical econometric models already hinted at the value of adaptive multi-horizon reasoning. The Heterogeneous Autoregressive model of realized volatility (HAR-RV)~\cite{corsi2009}, originally introduced for volatility forecasting, captures long-memory structure by linearly aggregating daily, weekly, and monthly rolling averages, demonstrating that multiple fixed look-back horizons carry complementary temporal information that no single-horizon model can subsume. HAR-RV and its descendants, however, commit to a designer-fixed set of horizons and to a linear functional form; the complementary horizons themselves, and the optimal way to combine them, remain hand-crafted. More recent deep-learning multi-period research, e.g. MLF~\cite{zhang25}, extend this multi-period intuition with sophisticated attention mechanisms and serves as our most competitive alternative. Nevertheless, MLF still relies on pre-defined periods and equally distributed patches. Such static designs prevent it from adaptively perceiving the optimal context in dynamic markets.

\begin{figure}[!t]
\centering
\begin{subfigure}[t]{1\linewidth}
    \includegraphics[width=\linewidth]{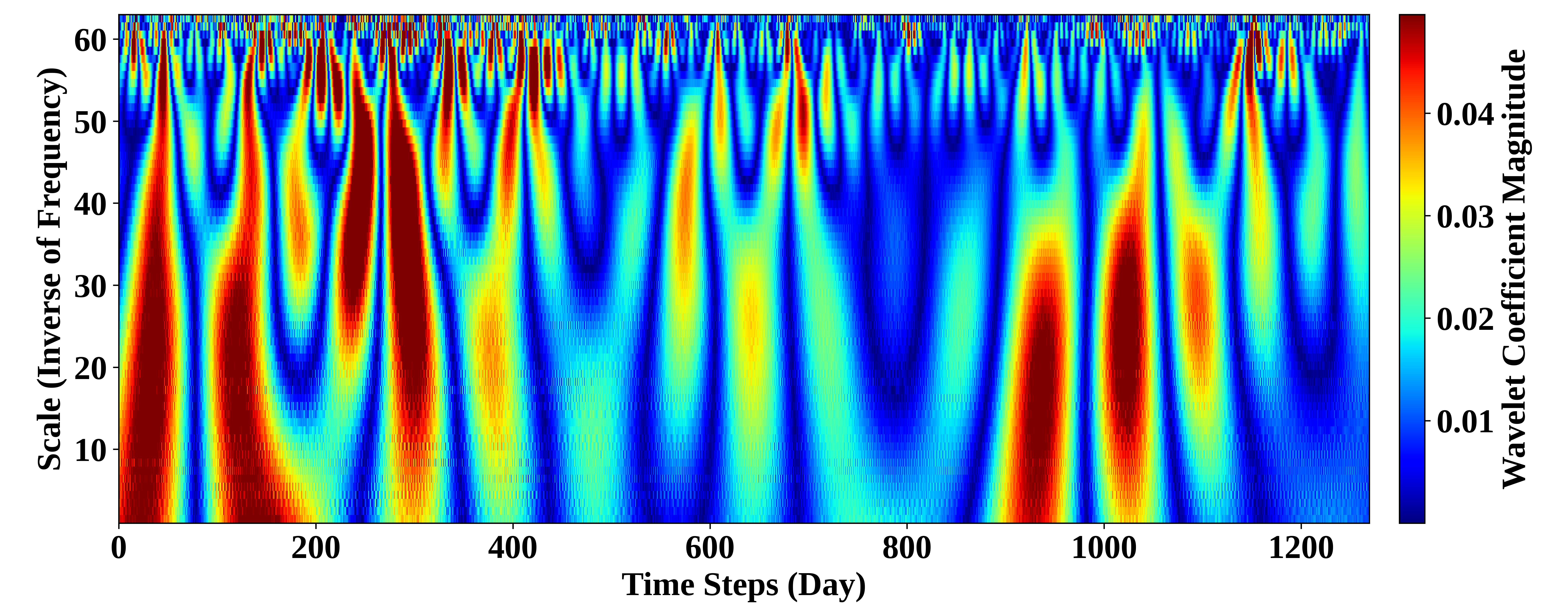}
    \caption{HS300 constituent 600176.SS (daily log-returns, 2020--2024)}
    \label{fig:cwt-fin}
\end{subfigure}
\vspace{6pt}\\
\begin{subfigure}[t]{1\linewidth}
    \includegraphics[width=\linewidth]{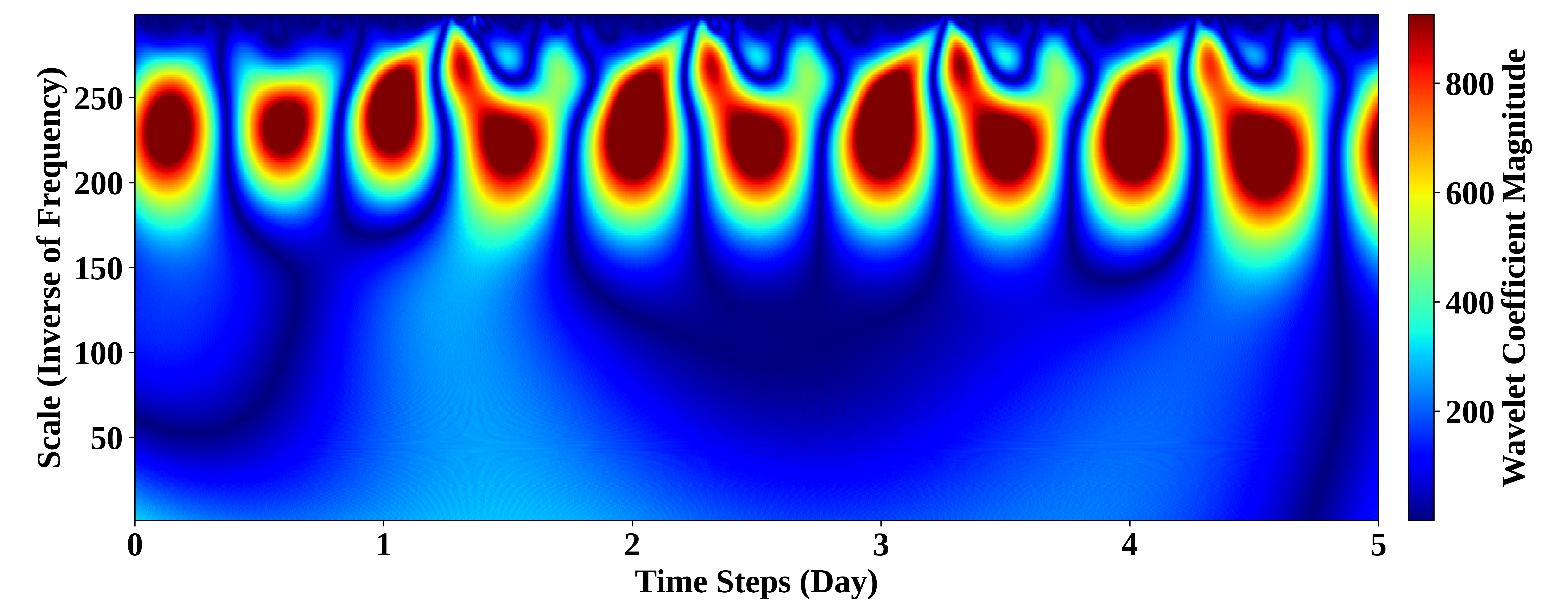}
    \caption{PEMS03 traffic flow (5-min)}
    \label{fig:cwt-pems}
\end{subfigure}
\caption{CWT scalograms for multi-scale analysis. Financial data (a) exhibits non-stationary energy distribution with no fixed periodicity, while traffic data (b) shows stable, periodic patterns.}
\label{fig:cwt}
\end{figure}

To empirically verify this claim, we apply the Continuous Wavelet Transform (CWT) as a multi-scale diagnostic:
\begin{equation*}
W_f(a,b) = \frac{1}{\sqrt{|a|}} \int_{-\infty}^{+\infty} f(t)\, \psi^*\!\left(\frac{t-b}{a}\right) dt,
\end{equation*}
where scale $a$ is inversely proportional to frequency and $b$ denotes temporal position. Figure~\ref{fig:cwt} visualizes the scalograms of two representative series. For the HS300 constituent 600176.SS (Figure~\ref{fig:cwt-fin}), energy concentration migrates unpredictably across scales within the five-year horizon. High-frequency components dominate in 2020, shift toward lower-frequency bands from 2021 to 2022, and return to high-frequency dominance by late 2023. No stable periodic structure persists at any scale. In contrast, the PEMS03 traffic series (Figure~\ref{fig:cwt-pems}) exhibits time-invariant energy bands at scales ranging from 200 to 260, reflecting a fixed daily periodicity that holds uniformly across the entire observation period. This divergence reveals that the dominant temporal scale governing predictive information in financial data is itself non-stationary. Thus, a fixed context window mechanism introduces an inductive bias that is mismatched to the underlying data-generating process.

To bridge this gap, we propose \textbf{\modelname} (Regime-Aware Variable-context Expert Network), a novel MoE-based framework designed for adaptive context modeling in financial time series forecasting. The core of \modelname lies in its learnable patch weighting and selection mechanism.
Unlike static methods that adopt a single fixed context length, \modelname dynamically evaluates the importance of each historical patch.
It accumulates these scores in reverse chronological order against Cumulative Importance Thresholding (CIT) based thresholds, and generates a nested sequence of consecutive look-back windows.
Each window is routed to a dedicated expert working at its corresponding temporal scale.
All windows are anchored at the most recent patch, ensuring temporal coherence of positional attention within each expert. To ensure that local specialization does not come at the cost of global coherence, we introduce a Global Compressed Representation (GCR) branch that runs in parallel over the full context. It distills a holistic global view that complements the local experts' selective, scale-specific processing. 
Furthermore, the nested routing topology creates structured overlap across expert inputs.
To address this issue, we propose the shape-aligned fusion with Correlation-Aware Weighting (CAW) strategy. It decorrelates expert representations prior to aggregation and eliminates redundant noise, yielding reliable multi-resolution forecasts.

Our main contributions are summarized as follows:
\begin{itemize}
\item 
\textbf{Dynamic Context Paradigm:} We identify the critical limitations of static, fixed-length context windows in non-stationary financial environments and propose \modelname. This framework adaptively adjusts the receptive field to time-varying market dynamics. It learns data-dependent look-back windows by accumulating patch importance in reverse order under CIT-based thresholds.
\item 
\textbf{Dual-View Architecture:} We design a dynamic MoE backbone augmented with a GCR branch.
The architecture balances local specialization and global context modeling.
Experts with distinct scales handle variable-length patches for fine-grained local perception.
Meanwhile, the GCR branch captures holistic historical information to preserve global coherence.
\item 
\textbf{Redundancy Mitigation Strategy:} We introduce Shape-Aligned Fusion and CAW. By dynamically compressing and decorrelating heterogeneous expert outputs, this strategy explicitly filters noise from overlapping input segments, enabling efficient utilization of MoE parameters under the low SNR of financial time series.
\item 
\textbf{Extensive Evaluation and Deployment progress:} 
We conduct extensive evaluations on cross-market cumulative log-return prediction. Compared with SOTA baselines, \modelname improves the Pearson correlation by 9.2\% on HS300 and 20.2\% on S\&P500, and reduces MSE by 18.2\% in fund sales forecasting. Cross-domain tests on four PEMS traffic datasets further verify its generalization ability, achieving best performance results across 14 out of 16 evaluated metrics. Under realistic backtest conditions, \modelname-driven strategies outperform our production baseline by over 10\% in cumulative returns, and the system is currently advancing through final online integration.

\end{itemize}

\section{\modelname Design}
\label{sec:method}

\begin{figure*}[!t]
\centering
\includegraphics[width=\linewidth]{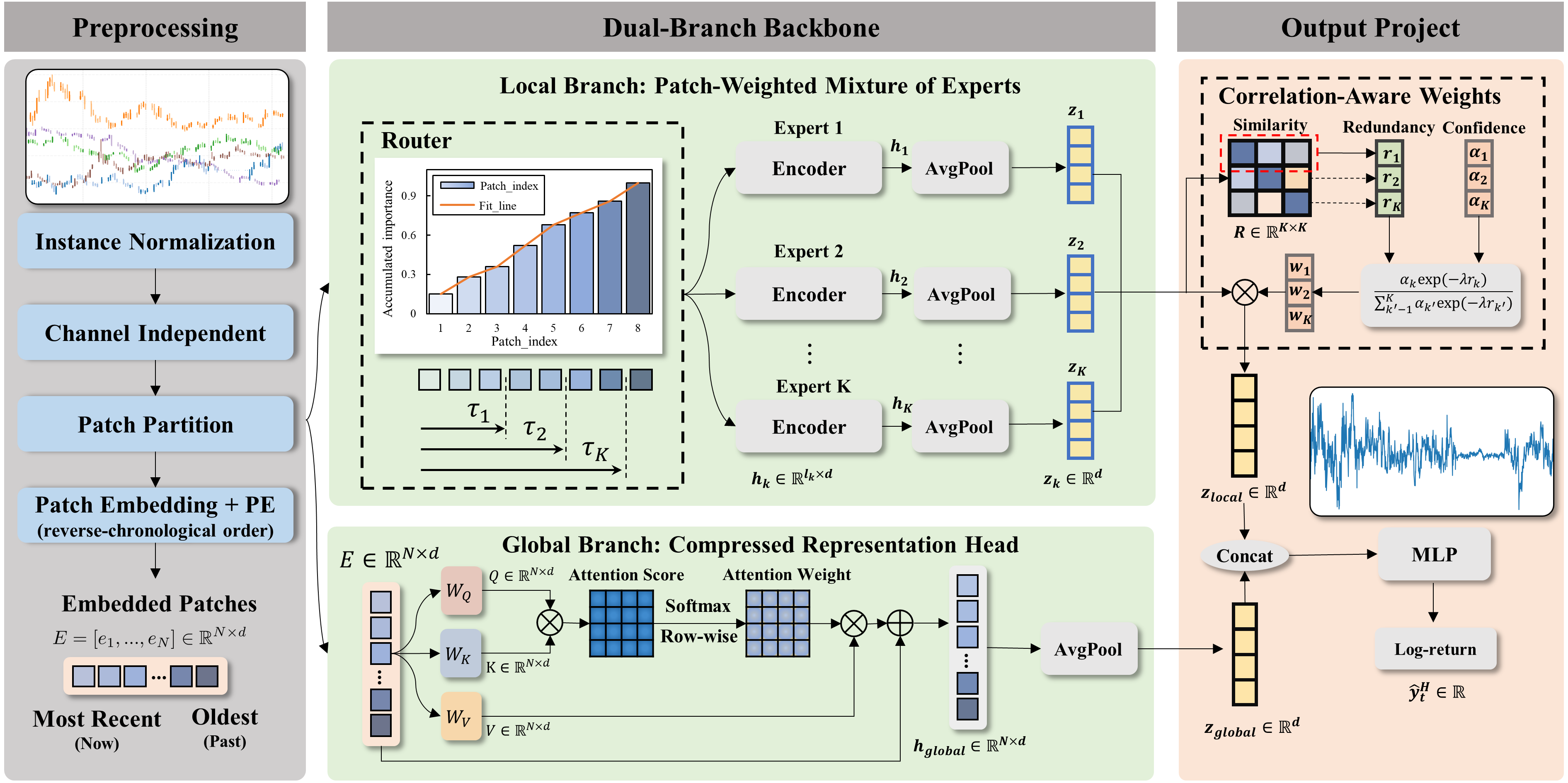}
\caption{Overview of \modelname. The pipeline consists of three modules. \textbf{Preprocess} applies instance normalization, channel-independent processing, and patch partitioning to produce embedded patches $\mathbf{E} = [\mathbf{e}_1, \ldots, \mathbf{e}_N]$. \textbf{Backbone} operates via two parallel branches. (i) The local adaptive branch scores patch importance and accumulates scores in reverse chronological order against CIT-based thresholds, generating $K$ nested contiguous look-back windows $\{\mathcal{G}_k\}$. Each window is processed by a scale-specialized expert, and the variable-length outputs are shape-aligned via average pooling into fixed-dimensional vectors for aggregation into $\mathbf{z}_{\text{local}}$. (ii) The GCR branch captures holistic historical dependencies across the full sequence $\mathbf{E}$ via a Self-Attention layer, then distills a global context vector $\mathbf{z}_{\text{global}}$ through average pooling. \textbf{Output Projection} concatenates $[\mathbf{z}_{\text{local}};\mathbf{z}_{\text{global}}]$ and projects them through an MLP head to output the final $H$-period cumulative log-return $\hat{y}_t^{(H)}$. The nested routing topology introduces collinearity among experts, which is jointly suppressed by the CAW scheme and the expert diversity regularizer.}
\label{fig:method}
\end{figure*}

\subsection{Problem Formulation}
\label{sec:method:problem}    

\begin{table}[!t]
\centering
\caption{Summary of notations.}
\label{tab:notations}
\vspace{-3pt}
\begin{tabular}{ll}
\toprule
\textbf{Notation} & \textbf{Explanation / Description} \\
\midrule
$D$ & Total dimension of the input channels \\
$L_{max}$ & Maximum historical look-back window length \\
$H$ & Forecast horizon for cumulative log return prediction \\
$B$ & Batch size \\
$X_t$ & Market state input matrix observed at time $t$ \\
$C_t$ & Asset closing price at time $t$ \\
$r_t$ & One-step log-return\\
$y_t^{(H)}$ & The $H$-period true cumulative log-return \\
$\hat{y}_t^{(H)}$ & The $H$-period forecasted log-return \\
\midrule
$p_{len}$ & Fixed patch length \\
$N$ & Total number of patches derived via $\lfloor L_{max}/p_{len}\rfloor$ \\
$d$ & Embedded feature dimension \\
$E$ & Joint patch embedding matrix \\
\midrule
$s_i$ & Raw importance score assigned to the $i$-th embedded patch \\
$\tilde{s}_i$ & Softmax-normalized patch importance \\
$K$ & Total number of experts \\
$\tau_k$ & The $k$-th cumulative sum threshold \\
$\Psi_i$ & Chronological reverse-cumulative patch importance score \\
$\mathcal{G}_k$ & Monotone patch index prefix subset to the $k$-th expert \\
$\ell_k$ & Dynamic selected patch number for expert $k$ \\
$h_k$ & Variable-length output hidden sequence \\
\midrule
$z_k$ & Dimension-aligned representation vector \\
$R$ & Expert cosine similarity to evaluate representation overlap \\
$r_k$ & Accumulated expert redundancy penalty score \\
$\alpha_k$ & Routing confidence of each expert \\
$w_k$ & Final adaptive gate weight \\
$\lambda$ & Learnable redundancy penalty\\ 
$z_{local}$ & Aggregated representation synthesized from local experts \\
$z_{global}$ & Compressed holistic feature vector from the global branch \\
\midrule
$\mathcal{L}_{MSE}$ & Log return forecasting loss utilizing mean squared error \\
$\mathcal{L}_{ent}$ & Router negative-entropy regularization penalty \\
$\mathcal{L}_{div}$ & Expert representation diversity regularizer \\
\bottomrule
\end{tabular}
\end{table}

We consider the task of multi-horizon return forecasting from multivariate financial time series. Let $\mathbf{x}_t \in \mathbb{R}^D$ denote the $D$-dimensional market state observed at time $t$, comprising standard market variables (e.g., OHLCV, bid-ask spread) and engineered factors. From the closing price $C_t$, we derive the one-step log-return:
\begin{equation}
r_t = \ln(C_t / C_{t-1}),
\label{eq:log_return}
\end{equation}
which encodes price innovations induced by the arrival of new market information \cite{hasbrouck91, gu20}.

\para{Input.} Given a maximum look-back length $L_{\max}$, an input instance at time $t$ is defined as
\begin{equation}
\mathbf{X}_t = [\mathbf{x}_{t-L_{\max}+1}, \dots, \mathbf{x}_t]^\top \in \mathbb{R}^{L_{\max} \times D}.
\label{eq:input_instance}
\end{equation}

\para{Target.} Given a forecast horizon $H$, the prediction target is the $H$-period cumulative log-return:
\begin{equation}
y_t^{(H)} = \sum_{h=1}^H r_{t+h} = \ln(C_{t+H} / C_t) \in \mathbb{R},
\label{eq:target}
\end{equation}
which corresponds to the realized holding-period return of a position entered at $t$ and liquidated at $t+H$.

\para{Objective.} The goal of \modelname is to learn a mapping $f_\theta: \mathbb{R}^{L_{\max} \times D} \to \mathbb{R}$, parameterized by $\theta$, such that the prediction $\hat{y}_t^{(H)} = f_\theta(\mathbf{X}_t)$ accurately approximates the true cumulative return (defined in Equation~\ref{eq:target}) over the out-of-sample test distribution~\cite{west96, diebold02}.

\subsection{Overall Architecture}
\label{sec:method:overview}

Figure~\ref{fig:method} illustrates the overall architecture of \textsc{\modelname}. Given a look-back window $\mathbf{X} \in \mathbb{R}^{L_{\max} \times D}$, the model produces a scalar $H$-period cumulative log-return forecast $\hat{y}_t^{(H)}$ through three functionally distinct stages:

\begin{itemize}[leftmargin=*, itemsep=2pt]
    \item \textbf{Preprocessing} (\S\ref{sec:method:preprocess}) normalizes the input instances, processes each channel independently, and partitions the look-back window into an embedded patch sequence $\mathbf{E} = [\mathbf{e}_1, \ldots, \mathbf{e}_N] \in \mathbb{R}^{N \times d}$.

    \item \textbf{Dual-Branch Backbone} (\S\ref{sec:method:backbone}) consumes $\mathbf{E}$ through two parallel pathways. The local branch routes nested contiguous patch prefixes to $K$ scale-specialized experts via a CIT-thresholded router. The global branch distills the entire look-back window into a single compressed representation.

    \item \textbf{Output Projection} (\S\ref{sec:method:output}) aligns the variable-length expert outputs to a common dimensionality, aggregates them via a CAW scheme, fuses the result with the global representation, and projects to the scalar forecast through an MLP head.
\end{itemize}

We elaborate on each module in the following sections.

\subsection{Preprocessing}
\label{sec:method:preprocess}

\noindent{\bf Instance normalization.}
We normalize each input instance to zero mean and unit variance over the temporal dimension of the look-back window. This removes sample-level distributional shift, a well-known obstacle in non-stationary financial series, while preserving the temporal structure of returns~\cite{Kim22}.

\para{Channel-independent processing.}
The normalized window $\mathbf{X} \in \mathbb{R}^{L_{\max} \times D}$ may contain heterogeneous channels, e.g., OHLCV bars, bid-ask spreads, engineered factors. Following~\cite{Nie23}, we process each channel independently through shared linear projections. This design prevents spurious information leakage across semantically distinct channels and allows the model to learn channel-specific temporal dynamics.

\para{Patch partitioning.}
We segment the look-back window into $N = \lfloor L_{\max} / p_{\text{len}} \rfloor$ non-overlapping patches $\{\mathbf{P}_1, \ldots, \mathbf{P}_N\}$~\cite{Nie23} arranged in reverse chronological order (most recent patch first). Each patch $\mathbf{P}_i \in \mathbb{R}^{p_{\text{len}}}$ is linearly projected to a $d$-dimensional embedding and augmented with a sinusoidal positional encoding:
\begin{equation}
    \mathbf{e}_i = \mathrm{PatchEmbed}(\mathbf{P}_i) + \mathrm{PE}(i) \in \mathbb{R}^d, i = 1, \ldots, N.
    \label{eq:patch-embedding}
\end{equation}

The resulting sequence $\mathbf{E} = [\mathbf{e}_1, \ldots, \mathbf{e}_N] \in \mathbb{R}^{N \times d}$ serves as the shared input to both branches of the backbone.

\subsection{Dual-Branch Backbone}
\label{sec:method:backbone}

The backbone comprises two parallel branches that jointly consume $\mathbf{E}$: a local branch that performs dynamic, scale-aware MoE routing over nested patch prefixes, and a global branch that maintains a holistic compressed view of the full look-back window. We describe each branch in turn.

\para{Local branch: Patch-weighted MoE.} The local branch comprises five steps.

{\em (i) Patch importance scoring.} A lightweight two-layer MLP assigns each embedded patch a scalar importance score $s_{i} = \phi(e_{i})$ for $i \in \{1, \dots, N\}$, where $\phi(\cdot)$ denotes the MLP scoring network with GeLU activation. These scores are then softmax-normalized over the sequence to yield a categorical distribution:
\begin{equation}
\tilde{s}_i = \frac{\exp(s_i)}{\sum_{j=1}^{N} \exp(s_j)}, \quad i = 1, \dots, N.
\end{equation}
Intuitively, $\tilde{s}_i$ reflects the model's learned belief about how informative patch $i$ is for the downstream forecast.

{\em (ii) CIT-thresholded routing.}
Unlike conventional MoE designs that route individual tokens to experts~\cite{Shazeer17,fedus22}, our router accumulates $\tilde{\mathbf{s}}$ from the most recent patch backward, constructing a reverse-chronological cumulative importance curve, and segments it at $K$ ordered thresholds $0 < \tau_1 < \tau_2 < \cdots < \tau_K \le 1$:
\begin{equation}
\mathcal{G}_k = \big\{\, i \in \{1,\dots,N\} : \Psi_i \le \tau_k \,\big\}, \quad \Psi_i = \sum_{j=1}^{i} \tilde{s}_j.
\label{eq:cit-groups}
\end{equation}
where \(k=1,\dots,K\).
By construction, $\mathcal{G}_1 \subseteq \mathcal{G}_2 \subseteq \cdots \subseteq \mathcal{G}_K$ forms a monotone chain of contiguous patch prefixes with data-dependent cardinalities $\ell_k = |\mathcal{G}_k|$, always anchored at the most recent patch. Because accumulation starts from the present and proceeds backward, each expert receives the $\ell_k$ most informative contiguous patches, a property critical for positional attention to remain temporally coherent. We set $K{=}3$, and the thresholds $(\tau_1, \tau_2, \tau_3) = (0.3, 0.6, 0.9)$ by default.

{\em (iii) Scale-specialized experts.}
Each group $\mathcal{G}_k$ is processed by an independent three-layer encoder $\mathrm{Encoder}_k$:
\begin{equation}
    \mathbf{h}_k = \mathrm{Encoder}_k\!\bigl(\{\mathbf{e}_i\}_{i \in \mathcal{G}_k}\bigr) \in \mathbb{R}^{\ell_k \times d}. 
    \label{eq:expert-output}
\end{equation}
Since $\mathcal{G}_1$ is the shortest (most recent) prefix and $\mathcal{G}_K$ the longest, expert $\mathrm{Encoder}_1$ naturally specializes in short-term dynamics while $\mathrm{Encoder}_k$ captures longer-term regime-scale structure, intermediate experts span intermediate horizons. This constitutes a fundamentally different form of specialization from content-based MoE routing: \textsc{\modelname} experts are differentiated by temporal scale, not by token content. Since Transformer encoders natively handle variable-length sequences, no padding is required within any expert.

{\em (iv) Shape-aligned pooling.}
To align the outputs from different scale-specialized experts into a unified space for downstream fusion, we apply a parameter-free Shape-aligned Pooling that averages over the patch dimension:
  \begin{equation}    
  \mathbf{z}_k = \text{AvgPool}(\mathbf{h}_k) \in\ \mathbb{R}^{d},
  \end{equation}
where $\text{AvgPool}(\cdot)$ averages over the $\ell_k$ patch embeddings in $\mathbf{h}_k \in \mathbb{R}^{\ell_k \times d}$, collapsing them into a fixed $d$-dimensional vector. This parameter-free design avoids overfitting risk and produces a uniform representation for the subsequent correlation-aware gate. 

{\em (v) CAW\_based expert aggregation.}
Because the expert groups are nested ($\mathcal{G}_1 \subseteq \cdots \subseteq \mathcal{G}_K$), their representations $\{\mathbf{z}_k\}$ are inherently correlated: longer-horizon experts operate on supersets of the inputs to shorter-horizon experts. A naive uniform or softmax gate would amplify this redundancy, which is particularly harmful under the characteristically low SNR of financial returns. To mitigate this, we compute the pairwise cosine similarity matrix $\mathbf{R} \in \mathbb{R}^{K \times K}$ with entries $\mathbf{R}_{jk} = \mathbf{z}_j^\top \mathbf{z}_k / (\|\mathbf{z}_j\|\|\mathbf{z}_k\|)$, derive the per-expert positive redundancy score $r_k = \sum_{j \neq k} \max(\mathbf{R}_{jk}, 0)$, and modulate a raw routing confidence $\alpha_k$ with an exponential penalty:
\begin{equation}
    w_k = \frac{\alpha_k \exp(-\lambda\, r_k)}{\sum_{k'=1}^{K} \alpha_{k'} \exp(-\lambda\, r_{k'})}, \qquad \lambda \ge 0 \;\text{(learnable)}.
    \label{eq:corr-aware-weights}
\end{equation}
The local representation is then obtained as the weighted aggregate:
\begin{equation}
    \mathbf{z}_{\text{local}} = \textstyle\sum_{k=1}^{K} w_k\, \mathbf{z}_k \in \mathbb{R}^d.
\end{equation}

\para{Global branch: GCR.}
To preserve a holistic macroeconomic view alongside fine-grained local experts, \modelname incorporates a GCR module. It first captures comprehensive historical dependencies across the entire sequence of embedded patches $\mathbf{E} \in \mathbb{R}^{N \times d}$ via a Self-Attention layer, then compresses the attended representations into a single vector through average pooling:
\begin{equation}
\mathbf{E}_{sa} = \text{Self-Attn}(\mathbf{E}),
\end{equation}
\begin{equation}
\mathbf{z}_{global} = \text{AvgPool}(\mathbf{E}_{sa}) \in \mathbb{R}^{d},
\end{equation}
where $\text{AvgPool}(\cdot)$ averages over the $N$ patch positions in $\mathbf{E}_{sa} \in \mathbb{R}^{N \times d}$, yielding a unified global context vector. The global branch thus serves as an information-preserving complement to the local branch's selective, scale-specialized view.

\subsection{Output Projection}
\label{sec:method:output}

The final prediction head fuses the two complementary representations via concatenation and projects to the scalar forecast:
\begin{equation}
    \hat{y}_t^{(H)} = \mathrm{MLP}\!\big(\mathrm{Concat}(\mathbf{z}_{\text{local}},\, \mathbf{z}_{\text{global}})\big) \in \mathbb{R}. 
    \label{eq:forecast-head}
\end{equation}
For multi-horizon forecasting, the output dimension of the MLP is extended to $M$, yielding one prediction $\hat{y}_t^{(H_m)}$ per target horizon $H_m$.

\subsection{Training Objective}
\label{sec:method:training}

We optimize a composite loss that combines the primary regression objective with two auxiliary regularizers targeting the specific failure modes of scale-aware MoE routing under low SNR:
\begin{equation}
    \mathcal{L} = \mathcal{L}_{\text{MSE}} + \lambda_{\text{ent}}\,\mathcal{L}_{\text{ent}} + \lambda_{\text{div}}\,\mathcal{L}_{\text{div}}.
    \label{eq:loss-total}
\end{equation}

\para{Forecasting loss.}
Consistent with the regression target defined in Eq.~\eqref{eq:target}, the primary objective is the mean squared error between the predicted and realized $H$-period cumulative log-return:
\begin{equation}
    \mathcal{L}_{\text{MSE}} = \frac{1}{B}\sum_{i=1}^{B}\big(\hat{y}_i^{(H)} - y_i^{(H)}\big)^2,
\end{equation}
where $B$ denotes the batch size.


\para{Router entropy regularization.}
The CIT-thresholded router operates on a softmax-normalized patch-importance distribution $\tilde{\mathbf{s}}$. Without regularization, the softmax parameterization tends to amplify initial score differences, concentrating all probability mass on a few patches, typically the most recent ones. This router collapse causes the cumulative sum to reach all thresholds $\{\tau_k\}$ within the first few patches, truncating every expert group into nearly identical short prefixes and degenerating the MoE toward a single-scale model. We counteract this with a negative-entropy penalty:
\begin{equation}
    \mathcal{L}_{\text{ent}} = \sum_{i=1}^{N} \tilde{s}_i \log \tilde{s}_i,
\end{equation}
which is minimized at the uniform distribution. A high-entropy routing distribution causes the cumulative importance curve to increase more gradually, yielding balanced prefix lengths across experts and enabling the intended short-, medium-, and long-horizon specialization.

\para{Expert diversity regularization.}
The nested structure $\mathcal{G}_1 \subseteq \cdots \subseteq \mathcal{G}_K$ means longer-horizon experts observe a strict superset of patches seen by shorter-horizon ones. Without explicit encouragement, experts may converge to similar representations (representation collapse), causing $\mathbf{z}_{\text{local}}$ to degenerate toward a single expert's output. We penalize the off-diagonal entries of the expert cosine-similarity matrix:
\begin{equation}
\begin{aligned}
\mathcal{L}_{\text{div}} &= \big\|\mathbf{R} - \mathbf{I}_K\big\|_F^2, \\
\mathbf{R}_{jk} &= \frac{\mathbf{z}_j^\top \mathbf{z}_k}{\|\mathbf{z}_j\|_2\|\mathbf{z}_k\|_2},
\end{aligned}
\label{eq:div_loss}
\end{equation}
which drives expert representations toward pairwise orthogonality. While the CAW (Eq.~\ref{eq:corr-aware-weights}) down-weights redundant experts at inference time, $\mathcal{L}_{\text{div}}$ structurally prevents redundancy during training. The two mechanisms are complementary.

\para{Complementary regularization.}
$\mathcal{L}_{\text{ent}}$ and $\mathcal{L}_{\text{div}}$ address orthogonal failure modes. The entropy term prevents routing collapse on the input side, ensuring every expert receives sufficiently rich context. The diversity term prevents representation collapse on the output side, ensuring experts produce non-redundant representations despite overlapping inputs. Their joint use is critical for stable training in the low-SNR financial regime (see ablation in \S\ref{sec:exp:ablation}).

\subsection{System Deployment}                                            
\label{sec:deployment}                                
    
\begin{figure}[!t]                                    
\centering
\includegraphics[width=\linewidth]{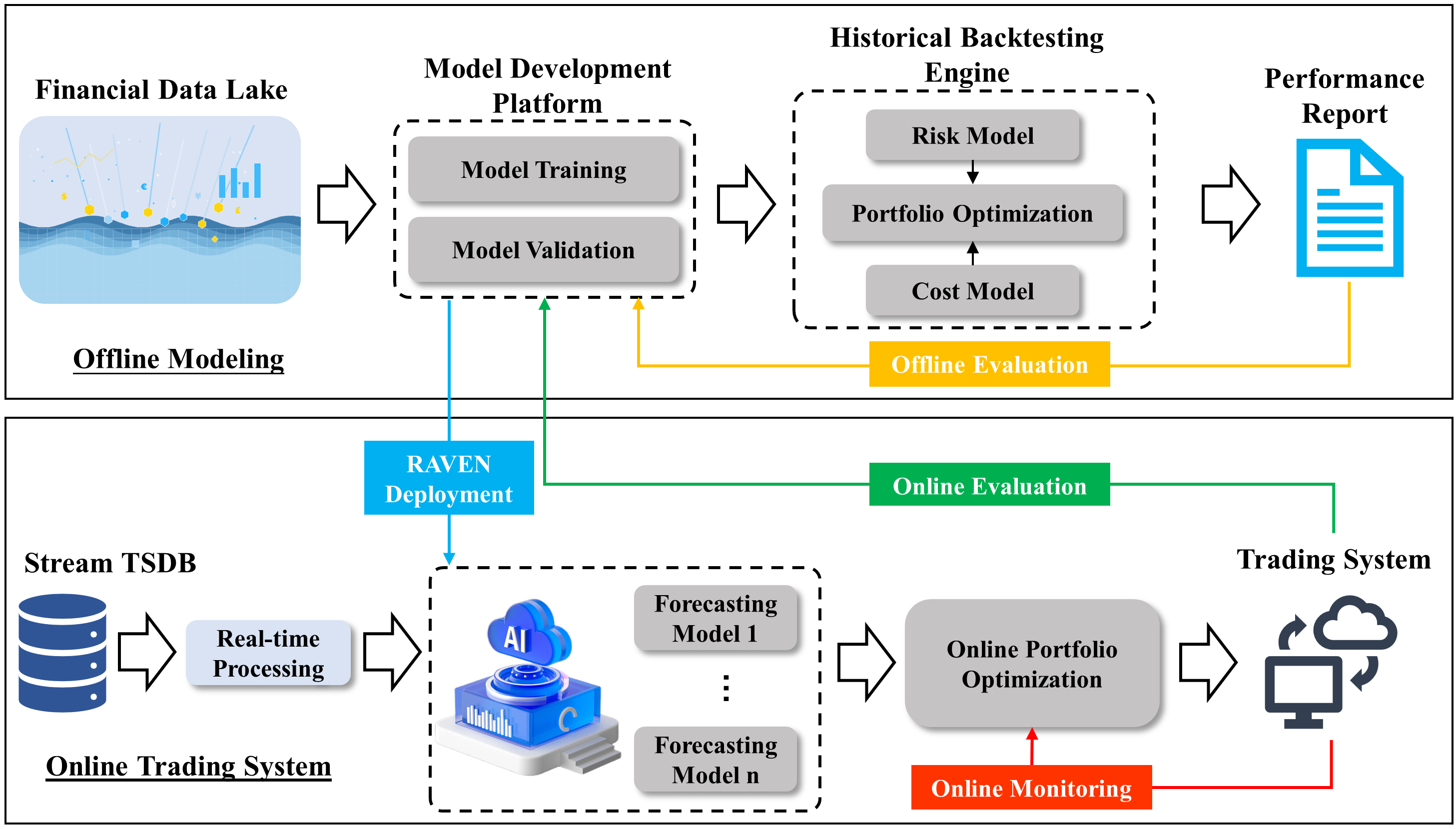}
\caption{Production deployment pipeline of \modelname in a quantitative trading system. The offline phase handles model training and backtesting validation on historical data. The online phase appends newly available market data after each close, generates return predictions via daily inference, optimizes portfolio allocations, and routes orders through pre-trade risk checks to execution venues. A production monitor triggers the next offline retraining cycle upon sustained performance drift.}
\label{fig:system}
\end{figure}

Figure~\ref{fig:system} illustrates the deployed system architecture, consisting of an offline development environment and an   
online production environment.                                              

\para{Offline environment.}
\modelname is trained on historical price-volume data stored in a financial data lake. The trained model then undergoes rigorous backtesting that incorporates portfolio optimization, risk constraints, and transaction cost modeling to validate out-of-sample performance. This offline cycle for model training is re-executed periodically, or when production monitoring signals significant performance degradation.

\para{Online environment.} Once deployed online, deployed forecasting models operates on a daily inference cycle. After each market close, newly available data is appended to the streaming database and fed into models to generate next-period return predictions across the target tickers. Predictions are consumed by a portfolio optimization module that solves for target allocation weights subject to risk and turnover constraints.
Optimized orders pass pre-trade risk checks at the trading console before routing to execution venues via broker bridges. A production monitor continuously tracks realized prediction accuracy and portfolio; sustained drift beyond predefined thresholds triggers the next offline retraining cycle.

\para{Deployment Status and Impact.}
At present, the offline modeling and backtesting validation phases have been completed. Under identical and realistic backtest conditions, \modelname-driven strategies successfully outperform our single production baseline by over 10\% in cumulative returns. The system is currently advancing through the final stages of live production integration, with active engineering efforts focused on online strategy ensembling and live out-of-sample monitoring prior to full-scale capital allocation.

\section{Experiments}
\label{sec:experiments}

We evaluate \modelname on financial log-return prediction and general time series forecasting tasks to demonstrate the effectiveness of dynamic look-back selection and correlation-aware expert aggregation.

\subsection{Experimental Settings}
\label{sec:exp_setting}

\noindent{\bf Datasets.} We conduct experiments on three categories of datasets. Table~\ref{tab:dataset_stats} summarizes the dataset statistics. 

(1)~\textbf{FinMultiTime}~\cite{Xu25}: A cross-market financial dataset containing daily price and volume data for 892 HS300 constituents and 4,694 S\&P~500 constituents, spanning from 2009 to 2024. This dataset covers diverse market regimes, including bull markets, bear markets, and high-volatility crisis periods across both Chinese and US equity markets.

(2)~\textbf{Fund}~\cite{zhang25}: Daily user transaction records for mutual fund subscriptions and redemptions on Alipay, spanning from January 2015 to January 2023. This dataset captures retail investor behavior dynamics and fund-level return patterns.

(3)~\textbf{PEMS} (03, 04, 07, 08)~\cite{Guo19}: Four widely used traffic flow datasets collected from the California Department of Transportation Performance Measurement System (PeMS). Each dataset records aggregated traffic flow measurements from highway sensor networks at 5-minute intervals. These datasets exhibit strong temporal non-stationarity due to rush-hour patterns, weekday/weekend shifts, and seasonal variations, serving to validate the generalizability of \modelname beyond financial domains.

\begin{table}[!t]
    \centering
    \caption{Dataset statistics. Ticker/Sensors denotes the number of stocks or funds in financial datasets or sensors in traffic datasets. Timepoints reports the total available data samples across all entities and time steps.}
    \label{tab:dataset_stats}
    \setlength{\tabcolsep}{2.8pt} 
    \begin{tabular}{llccc}
    \toprule
    \textbf{Domain} & \textbf{Dataset} & \textbf{Ticker/Sensors} & \textbf{Timepoints} & \textbf{Granularity} \\
    \midrule
    \multirow{3}{*}{\textbf{Financial}} 
    & HS300  & 892  & 18,836,466  & 1-Day \\
    & S\&P500 & 4694 & 104,692,416 & 1-Day \\
    & Fund     & 306  & 647,892     & 1-Day \\
    \midrule
    \multirow{4}{*}{\textbf{Traffic}} 
    & PEMS03   & 358  & 9,382,464  & 5-Min \\
    & PEMS04   & 307  & 15,649,632  & 5-Min \\
    & PEMS07   & 883  & 24,921,792 & 5-Min \\
    & PEMS08   & 170  & 9,106,560  & 5-Min \\
    \bottomrule
    \end{tabular}
\end{table}


\para{Baselines.}
We compare against twelve representative methods spanning diverse architectural strategies for temporal modeling:
\begin{itemize}[leftmargin=*, topsep=2pt, itemsep=1pt, parsep=1pt]
    \item \textbf{Static Patching models:} PatchTST~\cite{Nie23} operates on fixed uniform patches, while Patch-Concat and Patch\_Ensemble extend it via naive concatenation and ensembling over different lengths.
    \item \textbf{Predefined Multi-period models:} MLF~\cite{zhang25} utilizes explicitly predefined multi-period inputs combined with inter-period redundancy filtering.
    \item \textbf{Cross-variate interactions:} iTransformer~\cite{Liu24} and Crossformer~\cite{Zhang23} capture global inter-sensor dependencies through variate-as-token and cross-dimensional attention mechanisms.
    \item \textbf{Hierarchical Multi-resolution:} PathFormer~\cite{Chen24}, Scaleformer~\cite{Shabani23}, and NHits~\cite{Challu23} extract multi-scale representations internally through predefined hierarchical pathways, iterative refinement, or interpolation.
    \item \textbf{Frequency-domain transformations:} TimesNet~\cite{Wu23}, FEDformer~\cite{Zhou22}, and FiLM~\cite{Zhoumw22} tackle temporal dynamics via 2D-variations and spectral frequency decompositions.
\end{itemize}
All baselines are evaluated under identical data splits and preprocessing protocols for a fair comparison.

\para{Implementation details.}
\modelname uses $K{=}3$ experts with CIT-based thresholds $(\tau_1, \tau_2, \tau_3) = (0.3, 0.6, 0.9)$, patch length $p_{\text{len}} = 16$, and embedding dimension $d = 128$. Each expert is a 3-layer encoder. We optimize with AdamW for 60 epochs using cosine annealing. Auxiliary loss weights are $\lambda_{\text{ent}} = 0.1$ and $\lambda_{\text{div}} = 0.01$. All experiments run on a single NVIDIA A100 40G GPU.

\para{Evaluation metrics.}
For HS300 and S\&P500, we train on 2009--2019 and evaluate on 2020--2024 via rolling out-of-sample prediction of $H{=}10$ day cumulative log-returns. 
We report three complementary metrics: 
(1)~Pearson correlation (Corr) between predicted and realized returns, measuring directional accuracy; 
(2)~Mean Squared Error (MSE) of normalized predictions, capturing magnitude fidelity; and 
(3)~Information Coefficient Information Ratio (ICIR), defined as the mean of the cross-sectional IC divided by its standard deviation across rebalancing periods, quantifying the stability of the predictive signal. 
For per-year evaluation (Table~\ref{tab:performance_comparison_reduced}), all three metrics are computed within each calendar year. 
For the overall summary (Table~\ref{tab:finoverall}), Corr and ICIR are computed over the entire 5-year test period (2020--2024) to reflect long-horizon stability, while MSE is reported as the multi-year average. 
For Fund, we forecast fund sales at horizons $H \in \{1, 5, 8, 10\}$ days and report MSE and Weighted Mean Absolute Percentage Error (WMA for short). 
For general benchmarks, we adopt short-term horizons $H \in \{12, 24\}$ and report MSE and MAE.

\subsection{Financial Time Series Forecasting}
\label{sec:exp:financial}
Financial time series exhibit non-stationary dynamics and regime-dependent SNR ratios, making fixed look-back approaches particularly fragile. We evaluate \modelname on two distinct financial forecasting tasks: 10-day period stock cumulative log return prediction and fund sales forecasting.

\para{Results on FinMultiTime.}
Table~\ref{tab:performance_comparison_reduced} reports per-year results on HS300 and S\&P500. On HS300, \modelname ranks first or second in every year-metric cell, achieving 7 first-place and 8 second-place rankings out of 15. On S\&P500, \modelname achieves the highest Corr and ICIR across all 5 years, and the lowest MSE in 4 of 5 years, yielding 14 first-place rankings out of 15 cells. Overall, \modelname obtains the top rank in 21 out of 30 total year-metric-dataset cells (70\%), followed by MLF and NHits at 4 cells each. These results confirm that data-driven dynamic patch routing generalizes consistently across both datasets and all evaluation periods without manual scale configuration.

Table~\ref{tab:finoverall} summarizes the aggregated performance over the entire five-year test period from 2020 to 2024. On HS300, \modelname achieves a Corr of 0.0390, an ICIR of 0.3932, and an MSE of 0.9905. Compared with the second-best baseline, MLF, it improves Corr by 9.2\% and ICIR by 9.9\%, while reducing MSE by 0.27\%.
On S\&P500, \modelname attains a Corr of 0.0363, an ICIR of 0.5980, and an MSE of 1.001. It outperforms MLF by 20.2\% in Corr and 7.2\% in ICIR, while reducing MSE by 1.5\%. Across the two datasets, \modelname ranks first in all six dataset-metric pairs. The consistently higher ICIR suggests that \modelname produces not only stronger average predictions but also more stable predictive signals across rebalancing periods. This stability is particularly desirable for sequential decision-making systems that rely on robust ranking signals.

\begin{table*}[t]
\centering
\caption{Performance comparison on HS300 and S\&P500 (2020–2024). Corr and ICIR: higher is better ($\uparrow$); MSE: lower is better ($\downarrow$). Best in \textbf{bold}, second \underline{underlined}.}
\label{tab:performance_comparison_reduced}
\scriptsize
\renewcommand{\arraystretch}{1.2} 

\resizebox{\textwidth}{!}{%
\begin{tabular}{lccc|ccc|ccc|ccc|ccc}
\toprule
\multicolumn{16}{c}{\textbf{Dataset: HS300}} \\
\midrule
Model & \multicolumn{3}{c|}{2020} & \multicolumn{3}{c|}{2021} & \multicolumn{3}{c|}{2022} & \multicolumn{3}{c|}{2023} & \multicolumn{3}{c}{2024} \\
\cmidrule(lr){2-4} \cmidrule(lr){5-7} \cmidrule(lr){8-10} \cmidrule(lr){11-13} \cmidrule(lr){14-16}
 & Corr & MSE & ICIR & Corr & MSE & ICIR & Corr & MSE & ICIR & Corr & MSE & ICIR & Corr & MSE & ICIR \\
\midrule
\modelname & \textbf{.0567} & \underline{.9706} & \textbf{.5943} & \textbf{.0292} & \textbf{.9897} & \textbf{.1478} & \underline{.0422} & \underline{.9967} & \underline{.5224} & \textbf{.0300} & \textbf{.9978} & \underline{.3321} & \underline{.0500} & \underline{.9971} & \underline{.3765} \\
MLF & \underline{.0525} & \textbf{.9647} & \underline{.5457} & .0173 & 1.002 & .1056 & \textbf{.0583} & \textbf{.9786} & \textbf{.5467} & .0152 & \underline{1.002} & .2771 & .0388 & 1.004 & .3325 \\
NHits & .0338 & .9885 & .3222 & .0143 & \underline{.9943} & .1153 & .0238 & .9998 & .4677 & .0125 & 1.012 & .3147 & \textbf{.0525} & \textbf{.9938} & \textbf{.3772} \\
PathFormer & .0462 & .9759 & .5438 & \underline{.0216} & .9917 & \underline{.1363} & .0253 & .9996 & .4063 & \underline{.0176} & 1.003 & \textbf{.3332} & .0315 & 1.012 & .3080 \\
PatchTST & .0322 & .9893 & .3277 & .0043 & 1.012 & .0593 & .0193 & 1.006 & .4207 & .0089 & 1.012 & .2664 & .0429 & 1.001 & .3151 \\
Patch-Concat & .0373 & .9803 & .5082 & .0140 & 1.003 & .1190 & .0286 & .9974 & .4233 & .0138 & 1.005 & .2202 & .0428 & 1.002 & .3550 \\
Patch-Ensemble & .0320 & .9908 & .3477 & .0185 & 1.001 & .0756 & .0175 & 1.004 & .4469 & .0093 & 1.014 & .2657 & .0376 & 1.003 & .3092 \\
FiLM & .0372 & .9818 & .5282 & .0164 & 1.003 & .1123 & .0195 & 1.004 & .4709 & .0173 & 1.006 & .3280 & .0326 & 1.010 & .3200 \\
\bottomrule
\end{tabular}%
}

\vspace{3mm} 

\resizebox{\textwidth}{!}{%
\begin{tabular}{lccc|ccc|ccc|ccc|ccc}
\toprule
\multicolumn{16}{c}{\textbf{Dataset: S\&P500}} \\
\midrule
Model & \multicolumn{3}{c|}{2020} & \multicolumn{3}{c|}{2021} & \multicolumn{3}{c|}{2022} & \multicolumn{3}{c|}{2023} & \multicolumn{3}{c}{2024} \\
\cmidrule(lr){2-4} \cmidrule(lr){5-7} \cmidrule(lr){8-10} \cmidrule(lr){11-13} \cmidrule(lr){14-16}
 & Corr & MSE & ICIR & Corr & MSE & ICIR & Corr & MSE & ICIR & Corr & MSE & ICIR & Corr & MSE & ICIR \\
\midrule
\modelname & \textbf{.0339} & \textbf{1.003} & \textbf{.7256} & \textbf{.0365} & \textbf{.9989} & \textbf{.5892} & \textbf{.0426} & \underline{.9985} & \textbf{.6078} & \textbf{.0405} & \textbf{1.002} & \textbf{.5481} & \textbf{.0283} & \textbf{1.001} & \textbf{.5795} \\
MLF & .0268 & 1.027 & .6515 & .0283 & 1.023 & .5420 & .0350 & 1.011 & .5689 & .0339 & 1.008 & .5149 & .0207 & 1.013 & .5128 \\
NHits & .0268 & \underline{1.015} & .6423 & \underline{.0301} & \underline{1.001} & \underline{.5475} & \underline{.0379} & \textbf{.9957} & \underline{.5827} & .0310 & 1.007 & .5025 & .0221 & \underline{1.007} & .5143 \\
PathFormer & .0258 & 1.027 & .6337 & .0212 & 1.039 & .4737 & .0297 & 1.028 & .5204 & \underline{.0397} & \underline{1.006} & \underline{.5394} & .0220 & 1.009 & .5070 \\
PatchTST & .0195 & 1.042 & .5616 & .0219 & 1.032 & .5016 & .0225 & 1.037 & .4837 & .0305 & 1.014 & .4850 & .0126 & 1.026 & .4431 \\
Patch-Concat & .0242 & 1.027 & .6254 & .0262 & 1.023 & .5154 & .0234 & 1.032 & .5132 & .0324 & 1.011 & .5041 & .0186 & 1.013 & .4713 \\
Patch-Ensemble & \underline{.0302} & 1.017 & \underline{.6742} & .0275 & 1.021 & .5263 & .0314 & 1.021 & .5298 & .0342 & 1.007 & .5114 & \underline{.0243} & 1.009 & \underline{.5290} \\
FiLM & .0275 & 1.022 & .6524 & .0193 & 1.036 & .4724 & .0322 & 1.018 & .5511 & .0302 & 1.014 & .5091 & .0230 & 1.013 & .5127 \\
\bottomrule
\end{tabular}%
}

\end{table*}

\begin{table}[!t]
\centering
\caption{Overall average performance on HS300 and S\&P500 (2020--2024). Corr and ICIR: higher is better ($\uparrow$); MSE: lower is better ($\downarrow$). Best in \textbf{bold}, second \underline{underlined}.}
\label{tab:finoverall}
\setlength{\tabcolsep}{4.5pt}
\renewcommand{\arraystretch}{1.05}
\begin{tabular}{l|ccc|ccc}
\toprule
\multirow{2}{*}{Model} & \multicolumn{3}{c|}{HS300} & \multicolumn{3}{c}{S\&P500} \\
 & Corr  & MSE  & ICIR  & Corr  & MSE  & ICIR  \\
\midrule
\modelname & \textbf{.0390} & \textbf{.9905} & \textbf{.3932} & \textbf{.0363} & \textbf{1.001} & \textbf{.5980} \\
MLF & \underline{.0357} & \underline{.9932} & \underline{.3579} & \underline{.0302} & 1.016 & \underline{.5577} \\
NHits & .0282 & .9987 & .3210 & .0294 & \underline{1.002} & .5518 \\
PathFormer & .0278 & .9990 & .3260 & .0272 & 1.020 & .5300 \\
PatchTST & .0217 & 1.003 & .2758 & .0225 & 1.030 & .4836 \\
Patch-Concat & .0272 & .9987 & .3177 & .0254 & 1.023 & .5203 \\
Patch-Ensemble & .0218 & 1.003 & .2811 & .0290 & 1.018 & .5523 \\
FiLM & .0247 & 1.003 & .3463 & .0264 & 1.023 & .5360 \\
\bottomrule
\end{tabular}
\end{table}

\begin{figure}[!t]
\centering
\includegraphics[width=\linewidth]{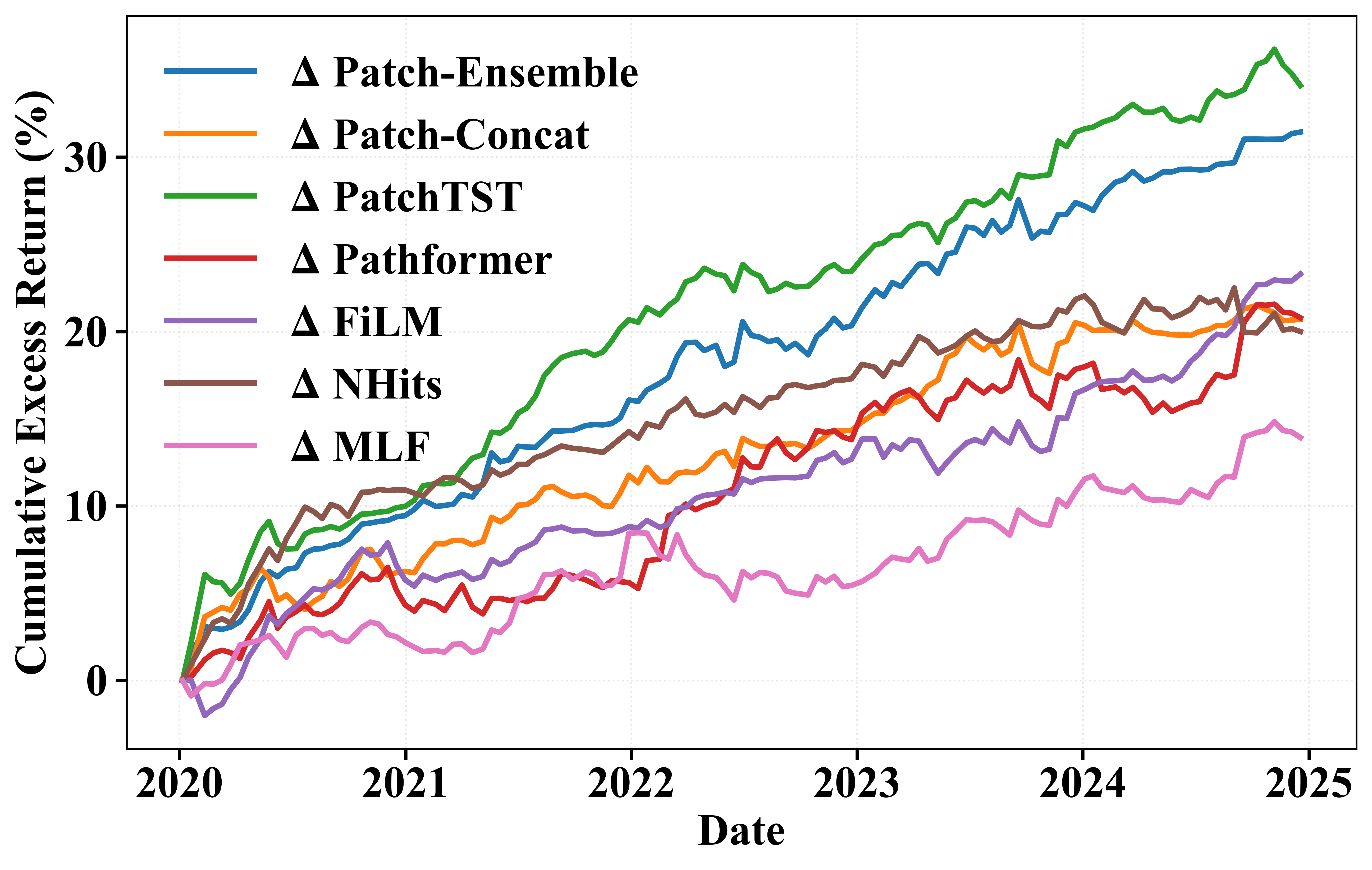}
\caption{Cumulative return advantage of \modelname over baselines on HS300, from 2020 to 2024. $\Delta(\cdot)$ denotes the cumulative return of \modelname minus that of baseline $(\cdot)$. All curves show a persistent upward trend across varying market regimes.}
\label{fig:backtest}
\end{figure}

To further validate that superior prediction accuracy translates into real-world investment gains, we conduct backtesting on HS300 from January 2020 to December 2024 using Qlib~\cite{yang20}. Qlib is an open-source AI-driven quantitative investment platform with standardized data pipelines and strategy simulation. We construct a cross-sectional portfolio management scheme. At each rebalancing point, i.e. every 10 trading days, constituents are ranked by predicted score in descending order, and a fixed-size portfolio of $K{=}30$ stocks is maintained. A drop-$N$ strategy ($N_{\text{drop}}{=}30$) limits turnover, as positions are only closed when their predicted rank falls significantly below the top-$K$ threshold. Execution is simulated via Qlib's SimulatorExecutor at daily frequency with an initial capital of 100M RMB and default transaction costs, including stamp duty and commission. We report the cumulative return of \modelname minus that of each baseline under the identical strategy and period.
Figure~\ref{fig:backtest} shows the cumulative return advantage of \modelname over each baseline. All gaps exhibit a persistent upward trend over the five-year period despite short-term fluctuations during volatile market phases, indicating sustained outperformance rather than episodic gains. The final cumulative return advantage over the strongest baseline (MLF) is +12.79\%, while the gap over PatchTST reaches +38.45\%. For the remaining baselines, the gaps are Patch-Ensemble (+34.41\%), FiLM (+23.41\%), PathFormer (+20.38\%), PathConcat (+20.28\%), and NHits (+19.46\%). The steadily growing gaps confirm that adaptive dynamic routing delivers consistent predictive advantage across both trending and volatile market regimes.

\para{Results on Fund.}
Table~\ref{table:fund} reports fund sales predictions across four horizons, where \modelname consistently achieves the best MSE and WMA. Specifically, \modelname reduces the average MSE by 18.2\% compared to the second-best MLF (32.42 vs. 39.62), maintaining stable gains from 14\% to 21\% across all horizons. Multi-scale architectures like PathFormer and Scaleformer perform similarly to single-scale methods, suggesting that implicit in-network multi-resolution modeling alone is insufficient. Explicit multi-period inputs with dynamic selection are imperative for capturing heterogeneous temporal patterns in transaction data. Moreover, fixed-window variants (Patch\_E, Patch\_C) consistently underperform MLF. This indicates that naively aggregating multi-period inputs without redundancy mitigation introduces correlated noise, negating the benefits of broader temporal context.

\begin{table*}[!t]
\centering
\caption{Performance comparison on Fund dataset. MSE ($\downarrow$) and WMA ($\downarrow$). Best in \textbf{bold}; second-best \underline{underlined}.}
\label{table:fund}
\vspace{-3pt}
\setlength{\tabcolsep}{3pt} 
\renewcommand{\arraystretch}{1} 
\begin{tabular}{c|cc|cc|cc|cc|cc|cc|cc|cc|cc}
\toprule
 & \multicolumn{2}{c|}{\modelname} & \multicolumn{2}{c|}{MLF} & \multicolumn{2}{c|}{PatchTST} & \multicolumn{2}{c|}{Patch-Ensemble} & \multicolumn{2}{c|}{Patch-Concat} & \multicolumn{2}{c|}{NHits} & \multicolumn{2}{c|}{FiLM} & \multicolumn{2}{c|}{Scaleformer} & \multicolumn{2}{c}{PathFormer} \\ \cmidrule{2-19} 
 & MSE & WMA & MSE & WMA & MSE & WMA & MSE & WMA & MSE & WMA & MSE & WMA & MSE & WMA & MSE & WMA & MSE & WMA \\ \midrule
1 & \textbf{27.69} & \textbf{74.63} & {\underline{33.85}} & {\underline{75.84}} & 36.68 & 81.05 & 40.3 & 86.68 & 39.68 & 85.87 & 36.53 & 80.29 & 46.12 & 96.1 & - & - & 35.54 & 80.75 \\
5 & \textbf{30.22} & \textbf{80.09} & {\underline{38.28}} & {\underline{80.37}} & 40.28 & 83.09 & 39.95 & 83.24 & 41.6 & 87.14 & 40.91 & 83.93 & 42.86 & 85.35 & 40.43 & 83.74 & 39.9 & 82.58 \\
8 & \textbf{33.62} & \textbf{85.32} & {\underline{41.94}} & {\underline{86.06}} & 44.71 & 88.81 & 44.06 & 88.77 & 43.49 & 88.08 & 44.81 & 89.23 & 44.57 & 89.12 & 43.97 & 87.62 & 44.17 & 88.67 \\
10 & \textbf{38.13} & \textbf{86.46} & {\underline{44.42}} & {\underline{88.66}} & 46.18 & 90.63 & 46.09 & 91.36 & 45.59 & 90.46 & 56.73 & 91.12 & 46.62 & 92.67 & 45.75 & 89.63 & 45.56 & 89.82 \\ \midrule
Avg. & \textbf{32.42} & \textbf{81.63} & {\underline{ 39.62}} & {\underline{ 82.73}} & 41.96 & 85.90 & 42.60 & 87.51 & 42.59 & 87.89 & 44.75 & 86.14 & 45.04 & 90.81 & 43.38 & 87.00 & 41.29 & 85.46 \\ \bottomrule
\end{tabular}
\end{table*}

\subsection{General Time Series Forecasting}
\label{sec:exp:general}
To evaluate the generalizability of \modelname beyond financial domains, we conduct experiments on four traffic flow datasets, namely PEMS03, PEMS04, PEMS07, and PEMS08~\cite{Guo19}. 
As shown in Table~\ref{tab:traffic_wide_correct}, \modelname achieves competitive results, securing the best performance in 14 of 16 evaluated metrics. The variate-as-token iTransformer emerges as the strongest second-best baseline, consistently outperforming the pre-defined multi-scale MLF. Traffic networks exhibit strict macroscopic spatial correlations, which allows iTransformer to excel by embedding historical series for global cross-sensor dependency capture. In contrast, MLF's static temporal squeezing inadvertently disrupts these spatial connections.

Despite the spatial dominance of traffic data, the overall superiority of \modelname reveals that predictive gains from data-dependent dynamic routing outweigh the lack of explicit spatial modeling. Traffic networks frequently suffer from abrupt, localized disruptions such as accidents or bottlenecks. iTransformer's static global receptive field risks over-smoothing these transient temporal bursts. Conversely, \modelname isolates these dynamics and adaptively assigns the optimal patch scale to each sensor's immediate state. This confirms that even in spatially-dominated domains, granular temporal adaptability provides a crucial inductive bias for forecasting volatile sequences.

\begin{table*}[t]
\centering
\caption{Short-term forecasting results on traffic datasets (MSE / MAE $\downarrow$). Best in \textbf{bold}; second-best \underline{underlined}.}
\label{tab:traffic_wide_correct}
\vspace{-3pt}
\resizebox{\textwidth}{!}{
\begin{tabular}{l | cccc | cccc | cccc | cccc}
\toprule
\multirow{3}{*}{\textbf{Model}} & \multicolumn{4}{c|}{\textbf{PEMS03}} & \multicolumn{4}{c|}{\textbf{PEMS04}} & \multicolumn{4}{c|}{\textbf{PEMS07}} & \multicolumn{4}{c}{\textbf{PEMS08}} \\
\cmidrule(lr){2-5} \cmidrule(lr){6-9} \cmidrule(lr){10-13} \cmidrule(lr){14-17}
& \multicolumn{2}{c}{$H=12$} & \multicolumn{2}{c|}{$H=24$} & \multicolumn{2}{c}{$H=12$} & \multicolumn{2}{c|}{$H=24$} & \multicolumn{2}{c}{$H=12$} & \multicolumn{2}{c|}{$H=24$} & \multicolumn{2}{c}{$H=12$} & \multicolumn{2}{c}{$H=24$} \\
\cmidrule(lr){2-3} \cmidrule(lr){4-5} \cmidrule(lr){6-7} \cmidrule(lr){8-9} \cmidrule(lr){10-11} \cmidrule(lr){12-13} \cmidrule(lr){14-15} \cmidrule(lr){16-17}
& MSE & MAE & MSE & MAE & MSE & MAE & MSE & MAE & MSE & MAE & MSE & MAE & MSE & MAE & MSE & MAE \\
\midrule
\modelname & \textbf{.067} & \textbf{.169} & \textbf{.091} & \underline{.204} & \textbf{.074} & \textbf{.172} & \textbf{.093} & \textbf{.195} & \underline{.072} & \textbf{.153} & \textbf{.086} & \textbf{.181} & \textbf{.077} & \textbf{.168} & \textbf{.110} & \textbf{.207} \\
MLF & .072 & .184 & .098 & .227 & .084 & .194 & .103 & .233 & .088 & .165 & .096 & .207 & .107 & .192 & .142 & .246 \\
PatchTST & .099 & .216 & .142 & .259 & .105 & .224 & .153 & .275 & .095 & .207 & .150 & .262 & .168 & .232 & .224 & .281 \\
iTransformer & \underline{.071} & \underline{.174} & \underline{.093} & \textbf{.201} & \underline{.078} & \underline{.183} & \underline{.095} & \underline{.205} & \textbf{.067} & .165 & \underline{.088} & \underline{.190} & \underline{.079} & \underline{.182} & \underline{.115} & \underline{.219} \\
NHits & .083 & .197 & .112 & .229 & .081 & .185 & .098 & .221 & .083 & \underline{.162} & .094 & .193 & .114 & .214 & .169 & .257 \\
TimesNet & .085 & .192 & .118 & .223 & .087 & .195 & .103 & .215 & .082 & .181 & .101 & .204 & .112 & .212 & .141 & .238 \\
FiLM & .089 & .208 & .120 & .233 & .079 & .188 & .106 & .227 & .092 & .182 & .117 & .226 & .124 & .225 & .195 & .273 \\
Patch-Ensemble & .107 & .239 & .159 & .285 & .118 & .247 & .176 & .290 & .104 & .221 & .153 & .267 & .182 & .246 & .231 & .298 \\
Patch-Concat & .103 & .224 & .131 & .247 & .096 & .218 & .157 & .279 & .094 & .213 & .143 & .255 & .149 & .228 & .192 & .260 \\
Crossformer & .090 & .203 & .121 & .240 & .098 & .218 & .131 & .256 & .094 & .200 & .139 & .247 & .165 & .214 & .215 & .260 \\
Scaleformer & .086 & .204 & .109 & .216 & .088 & .197 & .120 & .233 & .096 & .175 & .105 & .207 & .131 & .226 & .174 & .253 \\
PathFormer & .091 & .209 & .126 & .237 & .084 & .190 & .133 & .231 & .085 & .173 & .116 & .219 & .138 & .215 & .182 & .258 \\
FEDformer & .126 & .251 & .149 & .275 & .138 & .262 & .177 & .293 & .109 & .225 & .125 & .244 & .173 & .273 & .210 & .301 \\
\bottomrule
\end{tabular}
}
\end{table*}

\subsection{Ablation Study}
\label{sec:exp:ablation}

We ablate each component of \modelname on HS300 to isolate individual contributions. Table~\ref{tab:ablation} reports year-by-year and overall Pearson correlation for six variants.

\begin{table}[!t]
\centering
\caption{Ablation study on HS300 (Pearson Corr $\uparrow$). Best in \textbf{bold}.}
\label{tab:ablation}
\vspace{-3pt}
\setlength{\tabcolsep}{4pt} 
\renewcommand{\arraystretch}{1} 
\begin{tabular}{c|cccccc}
\toprule
Year & \textbf{\modelname} & w/o AR & w/o CAW & w/o GCR & w/o $\mathcal{L}_{\text{ent}}$ & w/o $\mathcal{L}_{\text{div}}$ \\
\midrule
2020 & 0.0567 & 0.0550 & 0.0569 & 0.0573 & 0.0583 & \textbf{0.0584} \\
2021 & \textbf{0.0292} & 0.0271 & 0.0260 & 0.0263 & 0.0290 & 0.0288 \\
2022 & \textbf{0.0422} & 0.0309 & 0.0313 & 0.0320 & 0.0321 & 0.0318 \\
2023 & \textbf{0.0300} & 0.0193 & 0.0184 & 0.0204 & 0.0202 & 0.0199 \\
2024 & \textbf{0.0500} & 0.0432 & 0.0441 & 0.0434 & 0.0449 & 0.0452 \\
\midrule
All  & \textbf{0.0390} & 0.0352 & 0.0358 & 0.0360 & 0.0373 & 0.0372 \\
\bottomrule
\end{tabular}%
\end{table}

Among the three architectural components, removing learned patch importance scoring without Adaptive Routing (AR) causes the largest degradation, with a 9.7\% relative drop on the overall metric. Without the learnable patch importance vector $\Psi_i$, the CIT mechanism degenerates into predefined window lengths, losing the ability to adapt look-back selection to regime-shifting dynamics. Disabling correlation-aware expert weighting (w/o CAW) degrades performance by 8.2\%, validating that explicit redundancy suppression at the output side is necessary when nested windows share overlapping patches. The global context representation (w/o GCR) contributes 7.7\%, providing a router-independent view that stabilizes predictions when local experts disagree.
The two auxiliary losses each contribute approximately 4.5\% individually. 


%

\subsection{Model Efficiency Analysis}
\begin{figure}[!t]
\centering
\includegraphics[width=\linewidth]{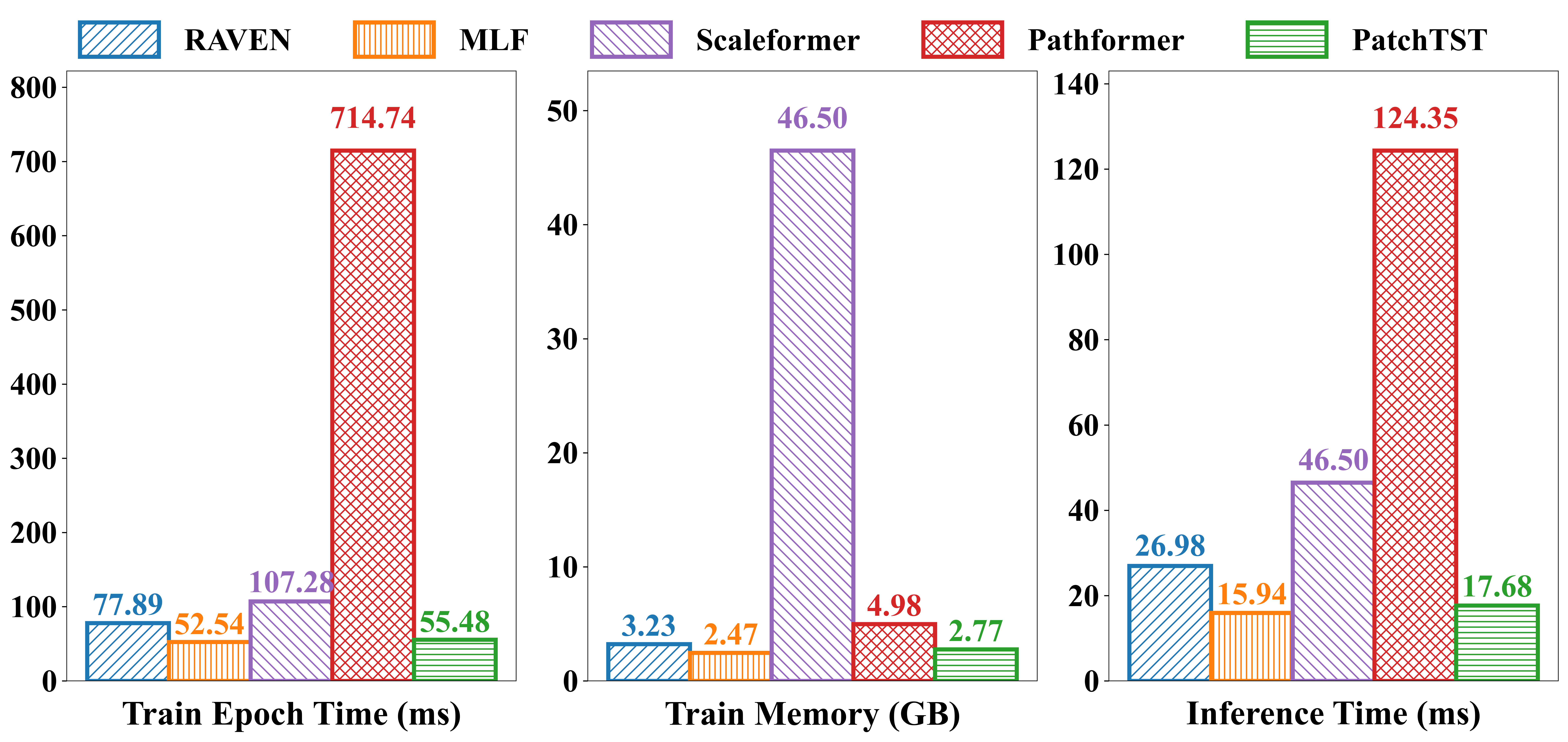}
\caption{Efficiency comparison on HS300 (batch size 512, look-back window 120). Training and inference time are reported in milliseconds per iteration; memory denotes peak GPU memory during training.}
\label{fig:efficiency}
\end{figure}

We benchmark the computational efficiency of \modelname against four representative baselines on HS300 using a single NVIDIA A100 (40GB) GPU, batch size 512, and a 120-day look-back window. Training and inference speeds are reported in milliseconds per iteration, where a training iteration denotes one full forward--backward pass over a batch and an inference iteration denotes one forward pass; memory denotes peak GPU memory during training. As shown in Figure~\ref{fig:efficiency}, \modelname achieves a favorable balance between accuracy and efficiency. Compared with lightweight baselines MLF and PatchTST, \modelname incurs moderate overhead in training time (77.89~ms vs. 52.54/55.48~ms), memory (3.23~GB vs. 2.47/2.77~GB), and inference latency (26.98~ms vs. 15.94/17.68~ms), reflecting the cost of adaptive temporal-scale routing. However, this overhead remains within the same order of magnitude, while \modelname achieves the best overall predictive performance on both HS300 and S\&P500 (Table~\ref{tab:finoverall}). Compared with heavier multi-scale or routing baselines, \modelname is substantially more efficient: it reduces training time by 27.4\% over Scaleformer and 89.1\% over PathFormer, reduces peak memory by 93.1\% over Scaleformer and 35.1\% over PathFormer, and reduces inference latency by 42.0\% over Scaleformer and 78.3\% over PathFormer. These results indicate that \modelname obtains the strongest predictive accuracy while maintaining computational cost close to lightweight baselines and far below heavier adaptive architectures.

\subsection{Hyperparameter Analysis}
\label{sec:exp:efficiency}
We analyze the sensitivity of \modelname to four key hyperparameters on HS300: number of experts $K$, CIT-based threshold values, maximum look-back window length, and patch length $p_{\text{len}}$. 

\para{Number of experts $K$.}
We vary $K \in \{2, 3, 4\}$ with thresholds uniformly spaced between $\tau_{\min}{=}0.3$ and $\tau_{\max}{=}0.9$. Table~\ref{tab:hyper-k} shows that $K{=}3$ achieves the best overall performance (0.0390), outperforming $K{=}2$ (0.0339) by 15.0\% and $K{=}4$ (0.0361) by 8.0\%. With $K{=}2$, only two scales are available, limiting the model's ability to capture intermediate-term patterns. With $K{=}4$, adjacent experts cover highly overlapping windows, increasing redundancy without providing additional discriminative information.

\begin{table}[!t]
\centering
\caption{Effect of number of experts $K$ on HS300 (Corr $\uparrow$).}
\label{tab:hyper-k}
\vspace{-3pt}
\begin{tabular}{c|ccc}
\toprule
Year & $K{=}2$ & $K{=}3$ & $K{=}4$ \\
     & (0.3, 0.9) & (0.3, 0.6, 0.9) & (0.3, 0.5, 0.7, 0.9) \\
\midrule
2020 & 0.0533 & 0.0567 & \textbf{0.0582} \\
2021 & 0.0273 & \textbf{0.0292} & 0.0274 \\
2022 & 0.0300 & \textbf{0.0422} & 0.0321 \\
2023 & 0.0173 & \textbf{0.0300} & 0.0184 \\
2024 & 0.0421 & \textbf{0.0500} & 0.0442 \\
\midrule
All  & 0.0339 & \textbf{0.0390} & 0.0361 \\
\bottomrule
\end{tabular}
\end{table}

\para{Threshold values under $K{=}3$.}
We compare three threshold configurations: (0.3, 0.6, 0.9), (0.2, 0.4, 0.8), and (0.1, 0.5, 0.9). As shown in Table~\ref{tab:hyper-tau}, all configurations yield comparable overall performance, with Corr ranging from 0.0370 to 0.0390 and less than 5.2\% relative variation. We further conduct paired $t$-tests on per-sample MSE using (0.3, 0.6, 0.9) as the reference. Neither (0.2, 0.4, 0.8) with $p{=}0.66$ nor (0.1, 0.5, 0.9) with $p{=}0.19$ shows a statistically significant difference at the 0.05 level. This insensitivity is anticipated. Since the importance scores $\tilde{s}_i$ are learned in an end-to-end manner, the model adjusts its score distribution to compensate for different threshold settings. Consequently, the exact threshold values are non-critical in practice.

\begin{table}[!t]
\centering
\caption{Effect of threshold values under $K{=}3$ on HS300 (Corr $\uparrow$).}
\label{tab:hyper-tau}
\vspace{-3pt}
\begin{tabular}{c|ccc}
\toprule
Year & (0.3, 0.6, 0.9) & (0.2, 0.4, 0.8) & (0.1, 0.5, 0.9) \\
\midrule
2020 & 0.0567 & \textbf{0.0602} & 0.0583 \\
2021 & 0.0292 & \textbf{0.0304} & 0.0303 \\
2022 & \textbf{0.0422} & 0.0313 & 0.0321 \\
2023 & \textbf{0.0300} & 0.0207 & 0.0199 \\
2024 & \textbf{0.0500} & 0.0442 & 0.0449 \\
\midrule
All  & \textbf{0.0390} & 0.0380 & 0.0370 \\
\bottomrule
\end{tabular}
\end{table}

\para{Maximum look-back window.}
As detailed in Table~\ref{tab:hyper-window}, performance improves steadily as the window extends from 60 days (0.0361) through 90 days (0.0378) to the optimal 120 days (0.0388). However, expanding the window further to 150 days leads to a slight degradation (0.0376). This inflection point suggests that excessively long historical periods introduce stale patterns that exacerbate concept drift and eventually surpass the filtering capacity of the CIT-based routing mechanism.

\begin{table}[!t]
\centering
\caption{Effect of maximum look-back window on HS300 (Corr $\uparrow$).}
\label{tab:hyper-window}
\vspace{-3pt}
\begin{tabular}{c|cccc}
\toprule
Year & 60 & 90 & 120 & 150 \\
\midrule
2020 & \textbf{0.0644} & 0.0585 & 0.0567 & 0.0584 \\
2021 & 0.0283 & 0.0290 & \textbf{0.0292} & 0.0289 \\
2022 & 0.0294 & 0.0322 & \textbf{0.0422} & 0.0330 \\
2023 & 0.0197 & 0.0198 & \textbf{0.0300} & 0.0204 \\
2024 & 0.0430 & \textbf{0.0458} & 0.0379 & 0.0452 \\
\midrule
All  & 0.0361 & 0.0378 & \textbf{0.0388} & 0.0376 \\
\bottomrule
\end{tabular}
\end{table}

\para{Patch length $p_{\text{len}}$.}
We vary $p_{\text{len}} \in \{8, 12, 16, 20\}$ to study the effect of routing granularity. As shown in Table~\ref{tab:hyper-plen}, \modelname is robust across the entire range, with overall Corr spanning from 0.0372 to 0.0390 (less than 5\% relative variation). Paired $t$-tests on per-sample MSE using $p_{\text{len}}{=}16$ as the reference confirm no statistically significant difference for any alternative ($p{=}0.26$, $0.21$, $0.18$ for $p_{\text{len}}{=}8$, $12$, $20$ respectively). This insensitivity arises because the learned importance scores $\tilde{s}_i$ adapt their distribution to the available routing granularity. With fewer patches, each score carries more discriminative weight, compensating for the coarser resolution.

\begin{table}[!t]
\centering
\caption{Effect of patch length $p_{\text{len}}$ on HS300 (Corr $\uparrow$). $N$ denotes the number of patches per window.}
\label{tab:hyper-plen}
\vspace{-3pt}
\begin{tabular}{c|cccc}
\toprule
Year & $p_{\text{len}}{=}8$ & $p_{\text{len}}{=}12$ & $p_{\text{len}}{=}16$ & $p_{\text{len}}{=}20$ \\
$N$  & 15 & 10 & 7 & 6 \\
\midrule
2020 & 0.0543 & \underline{0.0557} & \textbf{0.0567} & 0.0520 \\
2021 & \textbf{0.0303} & 0.0287 & \underline{0.0292} & 0.0271 \\
2022 & \underline{0.0415} & 0.0404 & \textbf{0.0422} & 0.0353 \\
2023 & \textbf{0.0315} & \underline{0.0314} & 0.0300 & 0.0289 \\
2024 & 0.0490 & 0.0473 & \textbf{0.0500} & \underline{0.0482} \\
\midrule
All  & 0.0384 & 0.0384 & \textbf{0.0390} & 0.0372 \\
\bottomrule
\end{tabular}
\end{table}

\subsection{Case Study}
\label{sec:exp:visualization}
\begin{figure}[!t]
\centering
\begin{subfigure}[b]{0.48\linewidth}
    \centering
    \includegraphics[width=\linewidth]{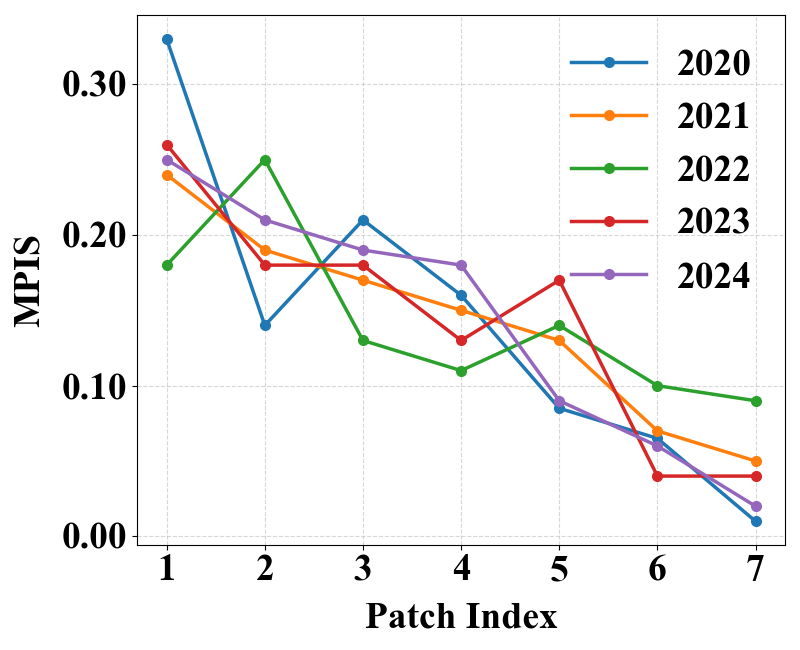}
    \caption{Yearly shift of 600176.SS.}
    \label{fig:patch_by_year}
\end{subfigure}
\hfill
\begin{subfigure}[b]{0.48\linewidth}
    \centering
    \includegraphics[width=\linewidth]{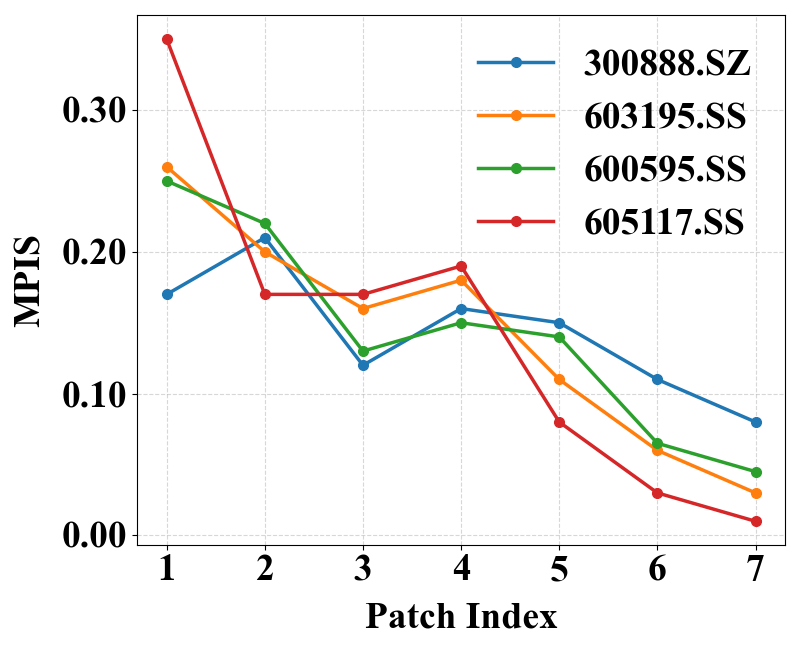}
    \caption{Cross-ticker profiles (2023).}
    \label{fig:patch_by_ticker}
\end{subfigure}
\caption{Distributions of Mean Patch Importance Score (MPIS) $\tilde{s}_i$ on HS300. Each data point represents the annual average of the learned importance at a given patch index. Patch index 1 corresponds to the most recent time segment. (a) Annual mean importance profiles of stock 600176.SS across five years, illustrating temporal regime adaptation. (b) Annual mean importance profiles of four stocks within 2023, illustrating cross-sectional heterogeneity.}
\label{fig:patch_importance}
\end{figure}

To understand how \modelname adapts its routing behavior, we visualize the learned patch importance scores $\tilde{s}_i$ under different conditions in Figure~\ref{fig:patch_importance}.

\para{Temporal heterogeneity.}
Figure~\ref{fig:patch_by_year} shows the annual mean importance profile of a single stock (600176.SS) across five years. In 2020, a year marked by extreme volatility, the router concentrates importance heavily on the most recent patch ($\tilde{s}_1 = 0.33$) with steep decay toward older patches, effectively selecting short look-back windows. In contrast, years with sustained trends such as 2021, 2023, and 2024 exhibit flatter distributions that allocate meaningful weight to intermediate and distant patches. This indicates that the router leverages longer historical context when the SNR of older observations is higher. The temporal shift emerges purely from data-driven learning of $\tilde{s}_i$, without explicit regime labels or calendar features.

\para{Cross-sectional heterogeneity.}
Figure~\ref{fig:patch_by_ticker} reveals that even within the same year (2023), different stocks exhibit markedly different annual mean importance profiles. Some stocks show steep concentration on recent patches, suggesting short-memory price dynamics, while others maintain more distributed weights across the full look-back range, benefiting from longer historical context. This cross-sectional variation demonstrates that no single fixed window length is universally optimal. Individual securities require different effective windows depending on their microstructure characteristics. \modelname addresses this by learning per-sample importance scores that drive CIT-based adaptive routing, enabling each input to select its own effective temporal scale without manual specification.

\begin{figure}[!t]
    \centering
    \begin{subfigure}{0.48\linewidth}
        \centering
        \includegraphics[width=\linewidth]{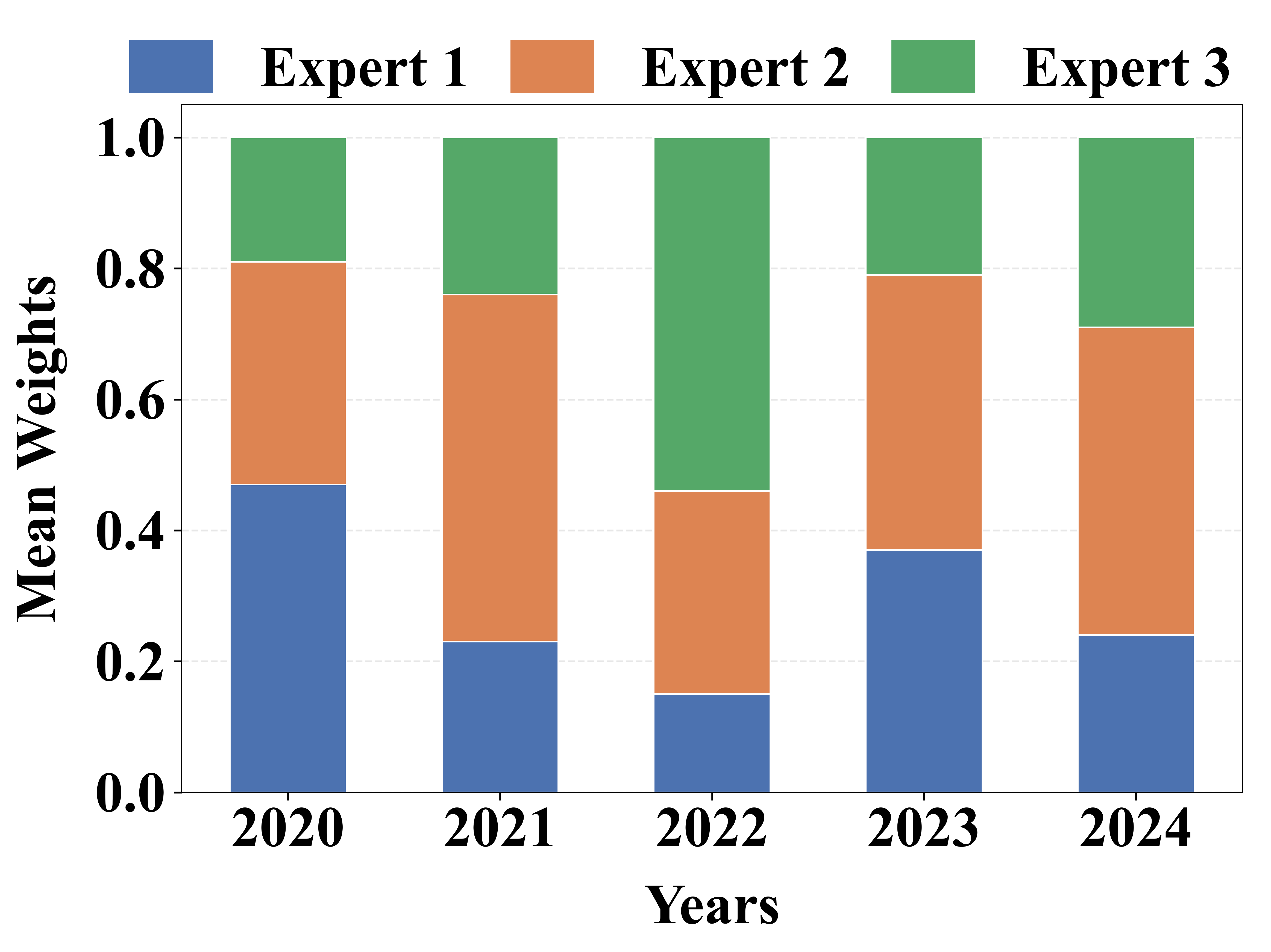}
        \caption{Temporal evolution of 605117.SS.}
        \label{fig:expert_year}
    \end{subfigure}
    \hfill
    \begin{subfigure}{0.48\linewidth}
        \centering
        \includegraphics[width=\linewidth]{
        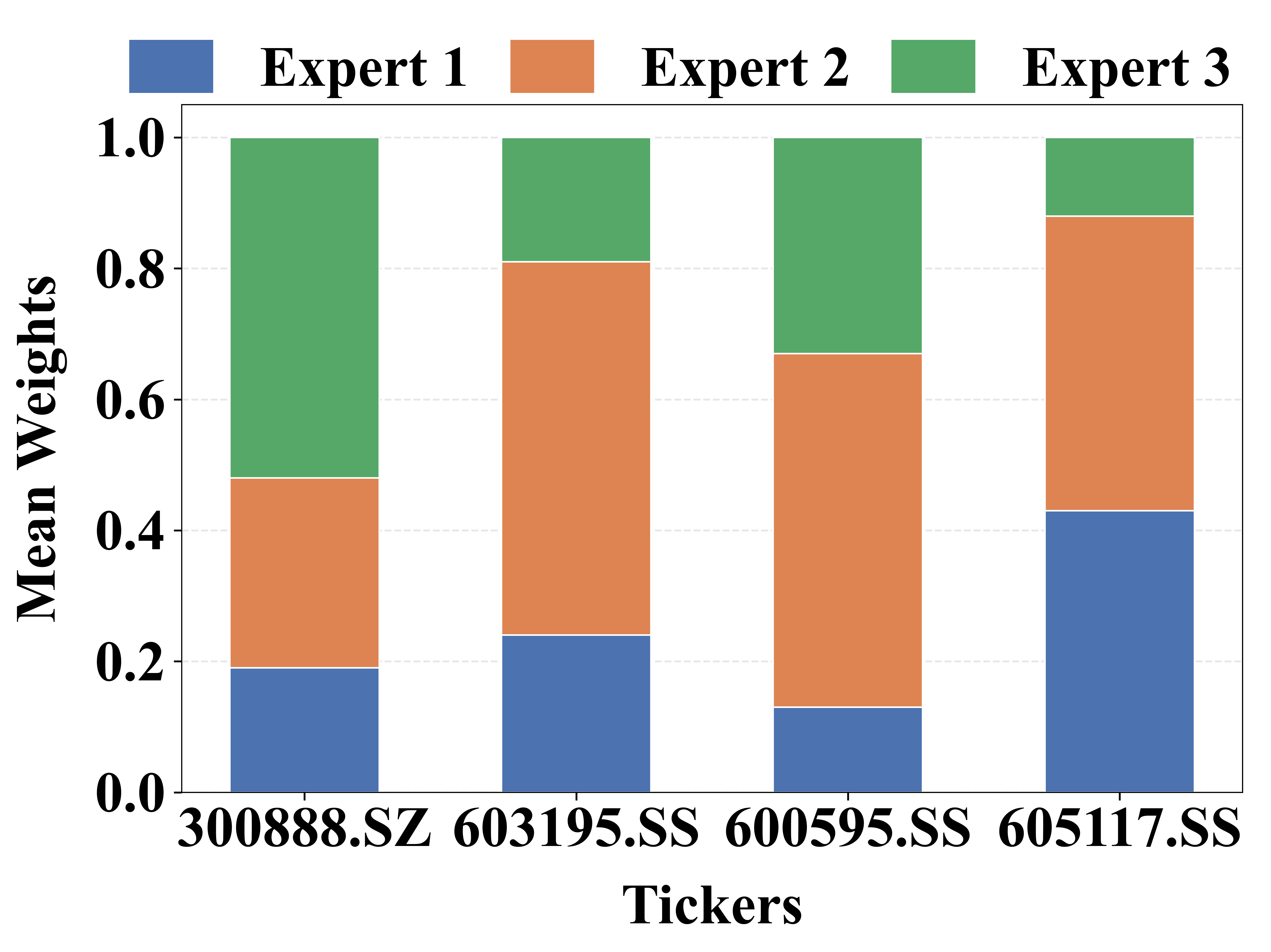}
        \caption{Cross-ticker distribution(2023).}
        \label{fig:expert_ticker}
    \end{subfigure}
    \caption{Empirical distributions of expert aggregation weights on HS300. Expert 1 corresponds to the short-horizon expert and Expert 3 to the long-horizon expert. Each bar shows the annual mean weight allocated to each expert. (a) Weight evolution of stock 605117.SS across five years, reflecting regime-driven reallocation. (b) Weight distribution across four stocks within 2023, reflecting asset-specific routing preferences.}
    \label{fig:expert_distributions}
\end{figure}

To establish a holistic understanding of \modelname's decision-making paradigm, we extend our visual inspection from the upstream patch importance curves to the downstream gate weights emitted at the expert aggregation stage. Figures~\ref{fig:expert_ticker} and \ref{fig:expert_year} illustrate the statistical distributions of the routing weights across different financial assets and historical periods, respectively. Crucially, these downstream weight topologies perfectly mirror the continuous context-carving behaviors observed in the upstream patch selection phase, confirming that the data-driven CIT-thresholded routing over learned importance matrices structurally translates into scale-specialized representations.

Figure~\ref{fig:expert_year} tracks the expert weight allocation of stock 605117.SS across five years. In 2020, a period of heightened volatility, Expert 1 (short-horizon) receives the largest share of weight, indicating that the model favors recent context when the market undergoes rapid structural change. In contrast, during 2021 and 2022, Expert 3 (long-horizon) dominates, as the model leverages extended historical context during relatively stable trending periods. The weight distribution shifts again in 2023 and 2024, reflecting changing market regimes. This adaptive reallocation confirms that the routing mechanism responds to non-stationary dynamics without manual intervention.

Figure~\ref{fig:expert_ticker} compares the weight allocations of four stocks within 2023. Stock 605117.SS allocates nearly half of its weight to Expert 1 (short-horizon), suggesting rapid price dynamics that benefit from short look-back contexts. In contrast, 300888.SZ assigns the majority of weight to Expert 3 (long-horizon), indicating stable temporal dependencies that reward extended historical input. The remaining two stocks fall between these extremes. This cross-sectional divergence mirrors the upstream patch importance heterogeneity observed in Figure~\ref{fig:patch_by_ticker}, validating that the entire routing pipeline maintains consistent behavior from patch scoring through expert aggregation.

\section{Related Work}

\noindent{\bf Deep learning and foundation models for financial forecasting.}
Classical financial forecasting relies heavily on gradient boosting trees (e.g., XGBoost~\cite{chen16}, LightGBM~\cite{ke17}), which capture non-linear interactions but treat forecasts as independent and identically distributed (i.i.d.) tabular problems, fundamentally discarding the temporal topology of market regimes~\cite{Bontempi12}. Subsequent sequential architectures, including RNNs~\cite{elman90} and LSTMs~\cite{hochreiter97}, restored temporal memory but remain disproportionately biased toward recency, frequently conflating transient microstructure noise with structural regime shifts~\cite{Sezer20}. 

To address long-range dependencies, general time-series architectures have evolved rapidly. Early efficient variants (Informer~\cite{Zhou21}, Autoformer~\cite{Wu21}) paved the way for advanced representation learning, precipitating breakthroughs in channel-independent patching (PatchTST~\cite{Nie23}), cross-variate dependencies (Crossformer~\cite{Zhang23}), inverted attention (iTransformer~\cite{Liu24}), and robust token blending (CARD~\cite{Wang24}). Complementary paradigms have also demonstrated strong competitiveness; notably, frequency-domain and wavelet models (FEDformer~\cite{Zhou22}, TimesNet~\cite{Wu23}, FredFormer~\cite{Piao24}, WPMixer~\cite{Murad25}) explicitly exploit spectral periodicity, while DLinear~\cite{Zeng23} and ModernTCN~\cite{Luo24} established robust baselines using linear decomposition and modernized temporal convolutions. 

In the highly stochastic financial domain, specialized architectures have emerged to capture complex market dynamics via graph-based trend prediction (HIST~\cite{xu21}), market-guided attention (MASTER~\cite{Li24}), meta-learning for distribution shifts (DoubleAdapt~\cite{Zhao23}), and early forms of conditional routing (TRA~\cite{Lin21}). 

However, despite their immense architectural diversity and expressive capacity, these models share a critical structural bottleneck: they strictly commit to a fixed context window or treat look-back length as a static hyperparameter. In the exceptionally low SNR ratio (SNR) and non-stationary environment of financial markets, a static receptive field inevitably forces models to either truncate critical historical regime transitions or dilute actionable predictive signals with obsolete noise.

\para{Adaptive context and non-stationarity in time series.}
In highly volatile and non-stationary time series domains, the temporal scale of a model's receptive field must adaptively expand or contract to isolate meaningful predictive signals from transient noise. The Multi-period Learning Framework (MLF)~\cite{zhang25} represents a notable attempt by simultaneously fusing varying look-back lengths using Inter-period Redundancy Filtering. Similarly, PathFormer~\cite{Chen24} captures varying temporal dynamics by dynamically aggregating features from a predefined set of fixed patch resolutions. More recently, TimeSqueeze~\cite{Ankireddy26} attempts to alleviate fixed-resolution bottlenecks by dynamically altering patch boundaries within a sequence based on local signal complexity. However, these models fundamentally commit to either a rigidly fixed global look-back window or a predefined set of static scales, failing to perceive the optimal global context length on the fly. \modelname bridges this gap at the architectural level. By deploying a data-driven CIT-thresholded routing mechanism over learned patch importance scores, \modelname dynamically generates nested, variable-length historical windows for each sample, seamlessly preserving continuous temporal evolution without human-defined scale constraints.

\para{MoE and routing topologies.}
Sparse MoE layers effectively scale model capacity without proportionally increasing computational overhead~\cite{fedus22}. In the time-series domain, foundation models like Time-MoE~\cite{Shi25} and Moirai-MoE~\cite{Liu25} scale this paradigm to billion parameters, employing token-level sparse routing. Recent specialized architectures such as TFPS~\cite{Sun25} follow a similar token-level paradigm, utilizing subspace clustering to route individual patches to pattern-specific experts. These content-based routing mechanisms are highly effective for localized pattern matching: experts specialize in patches with similar local morphology, regardless of where those patches appear in the sequence. However, such routing primarily induces pattern-level specialization rather than temporal-scale specialization. Since individual patches from different time positions may be dispatched to different experts, no expert is explicitly assigned to model a contiguous historical horizon. This limits the ability of each expert to capture how local patterns interact with medium- and long-range temporal context, which is important for non-stationary financial forecasting. \modelname adopts an orthogonal routing topology: temporal-scale routing. By ensuring each expert processes a contiguous, nested prefix of historical data, \modelname preserves positional coherence within each horizon while enforcing data-driven scale specialization.

\para{Positioning of our work.}
Unlike fixed-context models~\cite{Shi26, Ansari24} and predefined multi-scale methods~\cite{zhang25, Chen24, Ankireddy26}, \modelname performs sample-dependent context selection via CIT-thresholded routing; unlike token-level MoE~\cite{Shi25, Liu25, Sun25} that specializes experts by patch content, \modelname specializes by temporal scale through contiguous nested prefixes. Two additional components address challenges unique to this topology: (1)~Shape-Aligned Fusion with CAW aligns variable-length expert outputs and penalizes redundant representations caused by prefix overlap; (2)~a GCR branch summarizes the full context in parallel, providing macro-level information that localized experts may miss.

\section{Conclusion}
\label{sec:conclusion}

We presented \modelname, a regime-aware MoE framework that replaces fixed-length context windows with sample-adaptive, variable-length receptive fields for financial time series forecasting. By leveraging an importance-scoring mechanism coupled with CIT-based thresholds, \modelname dynamically extracts nested look-back windows routed to scale-specialized experts, while a GCR branch maintains macro-level context coherence. Furthermore, a CAW mechanism effectively decorrelates expert outputs, mitigating information redundancy stemming from overlapping patch inputs. 

Extensive experiments across financial and general time-series tasks, including cumulative log-return prediction on HS300 and S\&P500, fund flow forecasting, and traffic forecasting on PEMS benchmarks, demonstrate the effectiveness and generality of \modelname. On the two equity benchmarks, \modelname ranks first across all six overall metric--dataset combinations, improving Pearson correlation by 9.2\% on HS300 and 20.2\% on S\&P500 over the strongest baselines; on fund flow forecasting, it reduces MSE by 18.2\%. Beyond financial data, \modelname achieves the best result in 14 of 16 traffic forecasting metrics, indicating that adaptive temporal-scale routing is also beneficial for general time series. Ablation studies confirm that adaptive routing, CAW, and the GCR branch each contribute independently, while efficiency and hyperparameter analyses show that \modelname provides a favorable balance between accuracy and efficiency and remains stable across reasonable routing configurations.

While \modelname establishes a robust foundation for adaptive financial modeling, several promising avenues remain. First, integrating cross-asset dependencies and macro-regime indicators into the routing mechanism could better capture systemic market co-movements. Second, generalizing the framework to accommodate variable-resolution patching would further enhance its multi-scale expressiveness. Finally, scaling \modelname to high-frequency intraday data and broader asset classes (e.g., fixed income and commodities) represents a natural next step to validate its universal efficacy in complex financial ecosystems.

\clearpage

\bibliographystyle{IEEEtranS}
\bibliography{RAVEN}

\end{document}